%% file: main.tex
\documentclass[10pt,twocolumn,letterpaper]{article}

\usepackage{iccv}              

\definecolor{iccvblue}{rgb}{0.21,0.49,0.74}
\usepackage[pagebackref,breaklinks,colorlinks,allcolors=iccvblue]{hyperref}

\def\paperID{638} 
\def\confName{ICCV}
\def\confYear{2025}
\usepackage{url}
\usepackage{booktabs}       
\usepackage{amsfonts}       
\usepackage{nicefrac}       
\usepackage{enumitem}
\usepackage{microtype}      
\usepackage{xcolor}         
\usepackage{multirow,tabularx}
\usepackage{pifont}
\usepackage{graphicx}
\usepackage{amsmath}
\usepackage{amssymb}
\usepackage{algorithm}
\usepackage{wrapfig}
\usepackage{foreign}
\usepackage{multirow}
\usepackage{algorithm}
\usepackage{algpseudocode}
\usepackage{bbm}
\usepackage{mathtools}
\usepackage{xspace}
\usepackage{caption}

\makeatletter
\DeclareRobustCommand\onedot{\futurelet\@let@token\@onedot}
\def\@onedot{\ifx\@let@token.\else.\null\fi\xspace}
\def\eg{\emph{e.g}\onedot} 
\def\ie{\emph{i.e}\onedot}

\newcommand{\cmark}{\ding{51}}%
\newcommand{\xmark}{\ding{55}}%
\newcommand{\campose}{\pi}
\newcommand{\imagenotation}{X}
\newcommand{\depthnotation}{D}
\newcommand{\imagelatentnotation}{\mathbf{x}}  
\newcommand{\depthlatentnotation}{\mathbf{d}}
\newcommand{\layoutnotation}{\mathbf{l}}
\newcommand{\diffusionlatent}{\mathbf{z}}
\newcommand{\pointcloud}{\mathbf{p}}
\newcommand{\raynotation}{\mathbf{r}}
\newcommand{\categorynotation}{\mathbf{c}}
\newcommand{\positioninfonotation}{\mathbf{p}}
\newcommand{\novelviewnotation}{\mathbf{v}}

\makeatother

\title{HouseCrafter: Lifting Floorplans to 3D Scenes with 2D Diffusion Models}

\author{
Yiwen Chen$^{1}$, Hieu T. Nguyen$^{1}$,
 Vikram Voleti$^{2}$, Varun Jampani$^{2}$, Huaizu Jiang$^{1}$ \\
$^{1}$Northeastern University,~~~
$^{2}$Stability AI \\
{\tt\small \{chen.yiwe, nguyen.trungh, h.jiang\}@northeastern.edu, \{vikram.voleti, varun.jampani\}@stability.ai}
}

\begin{document}


\twocolumn[{
\renewcommand\twocolumn[1][]{#1}%
\maketitle
\centering
\includegraphics[width=0.9\textwidth]{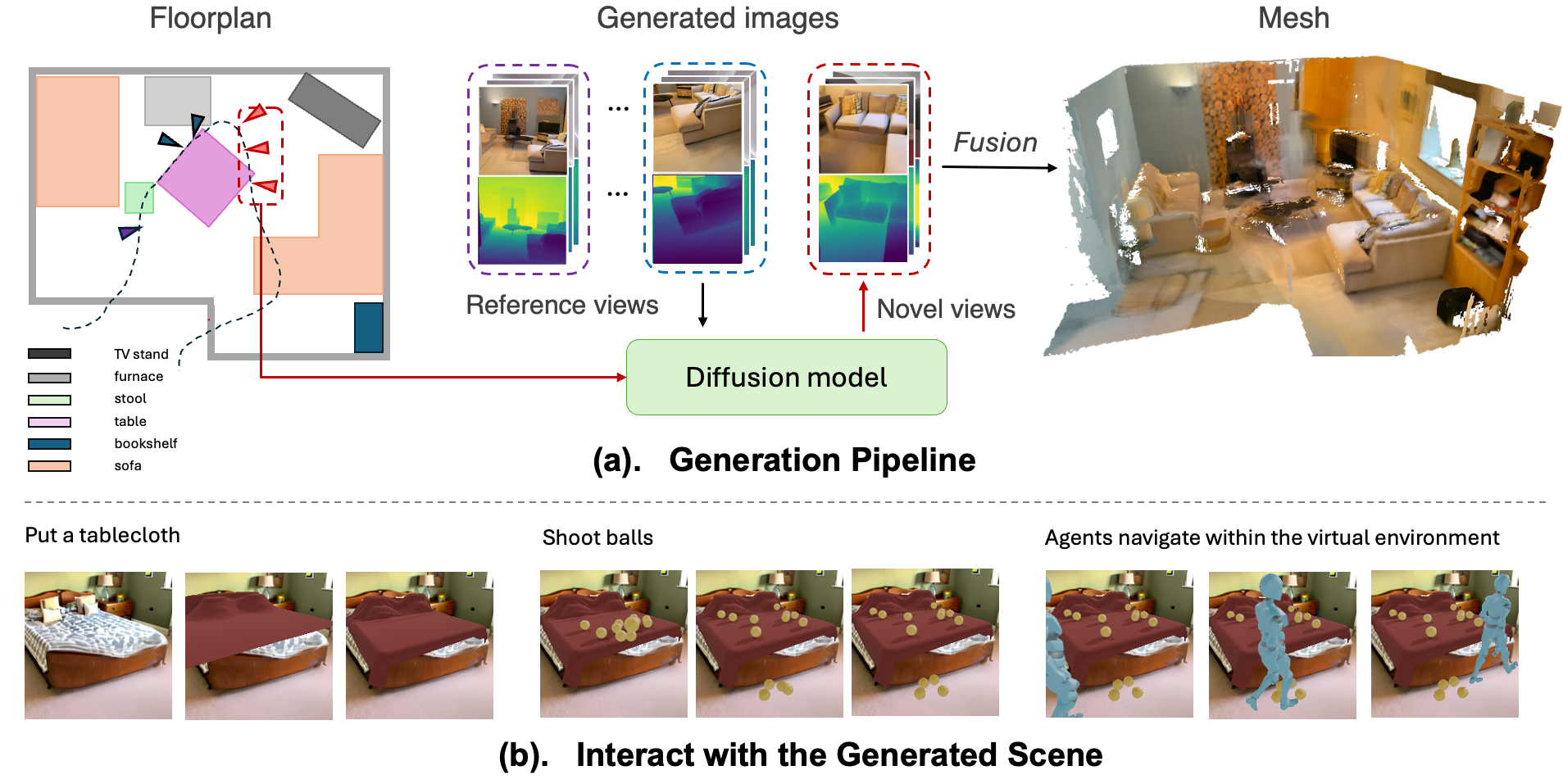}  
    \captionof{figure}{\textbf{Overview of HouseCrafter:} Our method lifts a 2D floorplan into a 3D scene using diffusion models, ensuring realistic geometry and structure.
    \textbf{Top:} Camera poses (triangles $\blacktriangle$) are sampled based on the floorplan.
    A 2D diffusion model is then used to generate RGB-D images batch-by-batch, where the generation of the $k$-th batch (\textcolor{red}{red}) is conditioned on the nearby poses (\textcolor{blue}{blue}) from the previous batch and the floorplan. The RGB-D images are then fused into a 3D mesh.
    \textbf{Bottom:} Our method generates a scene in explicit mesh form, which allows users to interact with the environment.
    }
    \label{fig:teaser}
    \vspace{10pt}
}]


\input{sec/0_abstract}    
\input{sec/1_intro}

\input{sec/2_RelatedWork}
\input{sec/3_Method}

\input{sec/4_Experiments}

\input{sec/5_Conclusion}
{
    \small
    \bibliographystyle{ieeenat_fullname}
    \bibliography{main}
}
\input{sec/Appendix}

\end{document}

%% file: sec/0_abstract.tex
\begin{abstract}
We introduce HouseCrafter, a novel approach that can lift a 2D floorplan into a complete large 3D indoor scene (\eg, a house). Our key insight is to adapt a 2D diffusion model, which is trained on web-scale images, to generate consistent multi-view color (RGB) and depth (D) images across different locations of the scene. Specifically, the RGB-D images are generated autoregressively in batches along sampled locations derived from the floorplan. At each step, the diffusion model conditions on previously generated images to produce new images at nearby locations. 
The global floorplan and attention design in the diffusion model ensures the consistency of the generated images, from which a 3D scene can be reconstructed. Through extensive evaluation on the 3D-FRONT dataset, we demonstrate that HouseCrafter can generate high-quality house-scale 3D scenes. Ablation studies also validate the effectiveness of different design choices. We will release our code and model weights.\end{abstract}

%% file: sec/1_intro.tex
\section{Introduction}

Interactable
3D environments are crucial for delivering truly immersive user experiences in AR, VR, gaming, and beyond. Typically, this process has been labor-intensive, demanding meticulous effort from skilled human artists and designers, especially for intricate indoor settings with numerous furniture pieces and decorative objects. The development of automated tools for generating realistic 3D scenes can significantly improve this process, streamlining the creation of complex virtual environments, which enables faster iteration cycles and empowers novice users to bring their creative visions to life. Such tools hold immense potential across industries like architecture, interior design, and real estate, facilitating rapid visualization, iteration, and collaborative design.

Recent advances in denoising diffusion models~\citep{ren2023xcube,ju2023diffroom} show great promise toward developing 3D generative models using 3D data.
In contrast to the abundant availability of 2D imagery~\citep{schuhmann2022laion}, 3D data requires intensive labor to create or acquire~\citep{dai2017scannet,chang2017matterport3d,front3d,ge2024behavior,behley2019semantickitti,yeshwanthliu2023scannetpp}.
Thus, using 2D generative models~\citep{rombach2022sd,saharia2022imagen} is a promising direction for 3D generation.
In~\citep{roomDreamer,MVDiffusion}, 2D diffusion models are used to texturize a given 3D scene with only the geometry. However, generating the untextured 3D scene as input for these methods is not trivial. Alternatively, 3D contents can be estimated based on generated multi-view observations~\citep{zero123, ye2023consistent,weng2023consistent123,syncDreamer, mvdream,shi2023zero123++,wonder3d,one2345,one2345p,kant2023invs,szymanowicz2023viewset,kant2024spad,wang2024crm, free3d, MVDFusion,huang2023epidiff,SV3D}.
However, the majority of existing works focus on investigating object-centric generation which has relatively simple camera positions and all images can be generated in one batch due to the small scale.
It is non-trivial to extend them for complex large-scale scene generation.

\textcolor{black}{
To address the challenge of 3D scene generation, many approaches leverage text-to-image diffusion models to synthesize indoor environments, either through image inpainting ~\citep{text2room,chung2023luciddreamer,shriram2024realmdreamer} or multi-view image generation~\citep{roomDreamer,MVDiffusion}. While these methods produce visually compelling results and enable language-driven control, they often struggle to generate large-scale scenes with accurate geometry and coherent object layouts. This limitation arises from their emphasis on local consistency rather than global scene structure, as well as the inherent imprecision of text prompts in specifying detailed spatial arrangements.
}
Instead of relying on textual descriptions, layout maps provide better global guidance for scene generation.
\textcolor{black}{
Several concurrent studies have investigated generating indoor environments conditioned on user-defined 3D layouts~\citep{ctrlRoom3d,ctrlRoom,bahmani2023cc3d,yang2024scenecraft}, utilizing box-shaped proxies to represent scene components and highlighting the advantages of incorporating layout information. However, as scene complexity and scale increase, managing 3D proxies becomes increasingly cumbersome for users. In comparison, 2D layout maps offer a more intuitive and efficient interface for specifying scene layouts.
}
\

In this paper, we present \textbf{HouseCrafter}, an autoregressive pipeline for \emph{house-scale} 3D scene generation guided by 2D floorplans, as shown in Fig.~\ref{fig:teaser}. 
Our key insight is to adapt a powerful pre-trained 2D diffusion model~\citep{rombach2022sd} to generate multi-view consistent images across different places of the scene in an autoregressive manner to reconstruct the 3D house. Specifically, we sample a set of camera poses within the scene based on the given floorplan. A novel view synthesis model is developed to generate images at these poses in a batch-wise manner. For each batch, the model takes the target poses and the already generated images at neighboring poses (initially empty) as reference to generates images at the target poses, guided by the local views of the floorplan. 
With all the generated images inside the house, we use the TSDF fusion~\citep{tsdf} to reconstruct the scene, providing explicit meshes for downstream applications (\eg, in an AR/VR application). With guidance from the floorplan, our method ensures global realism and consistency of images across batches, leading to high-quality scene generation.

Unlike existing novel view synthesis approaches ~\citep{escherNet, zero123, MVDFusion, syncDreamer}, our proposed model incorporates depth into both the reference and the novel/target views, where we consider both color and depth (RGB-D) images in the input and output.
This design choice offers two main advantages: (i) enhancing multi-view consistency within a single batch and across different batches in the autoregressive RGB-D image generation process and (ii) facilitating the final 3D scene reconstruction using the generated depth. Compared with previous approaches ~\citep{text2room, chung2023luciddreamer}, which suffer from depth scale ambiguity from monocular depth estimation models, our model outputs metric depth that can be directly used to reconstruct the scene. 
It is worth noting that RGB-D novel view synthesis has also been explored in object-centric novel view consistency~\citep{MVDFusion}. However, their approach focuses on generating low-resolution depth maps for better RGB view consistency. Instead, our approach generates high-resolution depth images for larges-scale scene reconstruction.

We evaluate our model on the 3D-Front dataset ~\citep{front3d}.
Through our experiments, we demonstrate the effectiveness of our RGB-D novel view synthesis model in generating images at the novel views that are consistent not only with the input reference views and floorplan, but also among the generated images themselves. 
Moreover, we demonstrate the model's efficacy in generating more compelling 3D scenes that are globally coherent than existing methods.

In summary, our key contributions are summarized as follows.
\begin{itemize}[itemsep=0.5pt,topsep=0pt,leftmargin=15pt]%
\item We introduce a novel method HouseCrafter, which can lift a 2D floorplan into a 3D house. Compared with other methods guided by \textcolor{black}{text or 2D layout infomation}~\citep{text2room,bahmani2023cc3d}, our approach can generate globally consistent house-scale scenes.
\item We present a RGB-D novel synthesis method, which takes nearby RGB-D images as reference to generate a set of RGB-D images at novel views, guided by the floorplan. 
Compared to existing RGB generation methods~\citep{escherNet,MVDFusion}, our approach generates semantically and geometrically consistent multi-view RGB-D images, enabling high-quality \textit{and efficient} 3D scene reconstruction.

\item Through both quantitative and qualitative evaluations, we demonstrate that our approach can generate globally coherent house-scale indoor scenes and faithful to the floorplan. Regarding the generated images, we demonstrate the effectiveness of our model in producing images that are faithful to both reference images and floorplan.
\end{itemize}

%% file: sec/2_RelatedWork.tex
\section{Related Work}
\label{sec:Related Work}

\noindent\textbf{3D Object Generation.}
Recent advancements in 2D image generation~\citep{rombach2022sd, stableVideo} have inspired attempts to use diffusion models for 3D generation. Some works~\citep{DreamFusion,magic3d,yi2023gaussiandreamer} optimize 3D representations~\citep{mildenhall2021nerf,kerbl3Dgaussians} by leveraging the denoising capabilities of diffusion models. However, these models struggle to maintain a single object instance across denoising updates and are unaware of camera poses, limiting the quality of the optimized 3D representations.

Alternatively, some works convert generated images into 3D models~\citep{zero123, ye2023consistent,weng2023consistent123,syncDreamer, mvdream,shi2023zero123++,wonder3d,one2345,one2345p,kant2023invs,szymanowicz2023viewset,kant2024spad,wang2024crm, tochilkin2024triposr,free3d, MVDFusion,huang2023epidiff}. \citet{zero123} demonstrated that diffusion models~\citep{rombach2022sd} fine-tuned on large-scale object datasets~\citep{deitke2023objaverse,deitke2024objaverseXL} can generate consistent multi-view RGB images, enabling 3D model reconstruction. Building on this, subsequent research has focused on enhancing multi-view image quality by integrating 3D representations ~\citep{consistnet,syncDreamer,kant2023invs,weng2023consistent123,mvdream,one2345,one2345p,MVDFusion} or using cross-view attention ~\citep{free3d,stableVideo,escherNet,mvdream,SV3D}.

Inspired by these approaches, we aim to generate multi-view images at the scene level. Our model uses multi-view RGB-D images and 2D floorplan as conditions to generate new multi-view RGB-D images. Integrating depth enhances multi-view consistency and provides explicit scene geometry for 3D reconstruction. Unlike\citet{escherNet}, which only outputs multi-view RGB images, and \citet{MVDFusion}, which denoises depth images with RGB latents, our model denoises both RGB and depth images in the latent space. This maintains geometry awareness and produces high-resolution depth images and high-quality 3D reconstructions, ensuring geometric and semantic consistency across views.

\noindent\textbf{Text-guided 3D Scene Generation.} 
Text-to-image models can be also utilized for 3D scene generation.
Some works ~\citep{pixelSynth,text2nerf,yu2023wonderjourney,chung2023luciddreamer,ouyang2023text2immersion,text2room,shriram2024realmdreamer} 
continuously aggregates frames with existing scenes, using monocular depth estimators to project 2D images into 3D space, but faces challenges like scale ambiguity and depth inconsistencies.
Recent work improves geometry by training depth-completion models~\citep{engstler2024invisible}. However, most of these methods focus on forward-facing scenes, struggling for larger or holistic scenes like entire rooms or houses since global plausibility is not guaranteed ~\citep{text2room}.

To enhance global plausibility, MVDiffusion ~\citep{MVDiffusion} and Roomdreamer ~\citep{roomDreamer} generate multiple images in a single batch to form a panorama, though without geometry generation. Gaudi~\citep{bautista2022gaudi}, directly generates global 3D scene representation, producing 3D scenes with globally plausible content, but the quality is limited by the scarcity of 3D data with text.

Inspired previous works, our pipeline generates views of the scene autoregressively but in batches. Compared to image-by-image generation pipelines~\citep{text2room,chung2023luciddreamer,shriram2024realmdreamer}, batch generation scales better and benefits from the built-in cross-view consistency of multi-view models. Additionally, by including depth images, HouseCrafter addresses scale ambiguity and leverages geometry from previous steps to generate novel views.

\noindent\textbf{Layout-guided 3D Scene Generation.} 
Complimenting to text, the layout provides the detailed position of objects in the scene. Early work~\citep{vidanapathirana2021plan2scene} is able to uplift a 2D floorplan to a 3D house model but only focuses on the architectural structure, \ie floor, wall, ceiling. Also conditioned on 2D layout, BlockFusion ~\citep{blockfusion} achieves commendable results in geometry generation but does not generate texture.

For joint geometry and texture generation, Ctrl-Room~\citep{ctrlRoom} and ControlRoom3D~\citep{ctrlRoom3d} demonstrate that incorporating 3D layout guidance to generate a room significantly improves geometric fidelity and object arrangement compared to text-only methods~\citep{text2room}.
\textcolor{black}{
SceneCraft~\citep{yang2024scenecraft} further extends this idea by introducing a framework for generating house-scale scenes guided by 3D bounding boxes of scene components. However, relying on 3D proxy boxes for guidance requires that users spend more effort managing and designing the proxies. In contrast, representing the scene with a 2D layout can provides a more user-friendly alternative, as it simplifies the editing of scene composition.
}
CC3D~\citep{bahmani2023cc3d}, closest to our work, uses 2D layout guidance to produce a 3D neural radiance field, enabling textured mesh but still limited to single-room scenes. Our method effectively uses 2D layout guidance to scale to larger scenes, such as entire houses.

\textbf{Other works.}
Other approaches treat indoor scene generation as an object layout problem~\citep{,wen2023anyhome,feng2024layoutgpt,yang2024holodeck,ATISS,Nie_2023_CVPR,shabani2022housediffusion,echoscene,tang2024diffuscene}. These works focus on predicting floor layouts and furniture placement using with language model, and retrieving suitable objects from a database. Alternatively, \citet{ge2024behavior} create augmented layouts from templates, while others use procedure generation~\citep{procThor,raistrick2024infinigen}   These approaches complement our pipeline, as we can use predicted floorplans to generate the scene’s texture and geometry accordingly.

%% file: sec/3_Method.tex
\begin{figure}[t!]
  \centering
  \graphicspath{{img/}}
  \includegraphics[width = 0.95\linewidth]{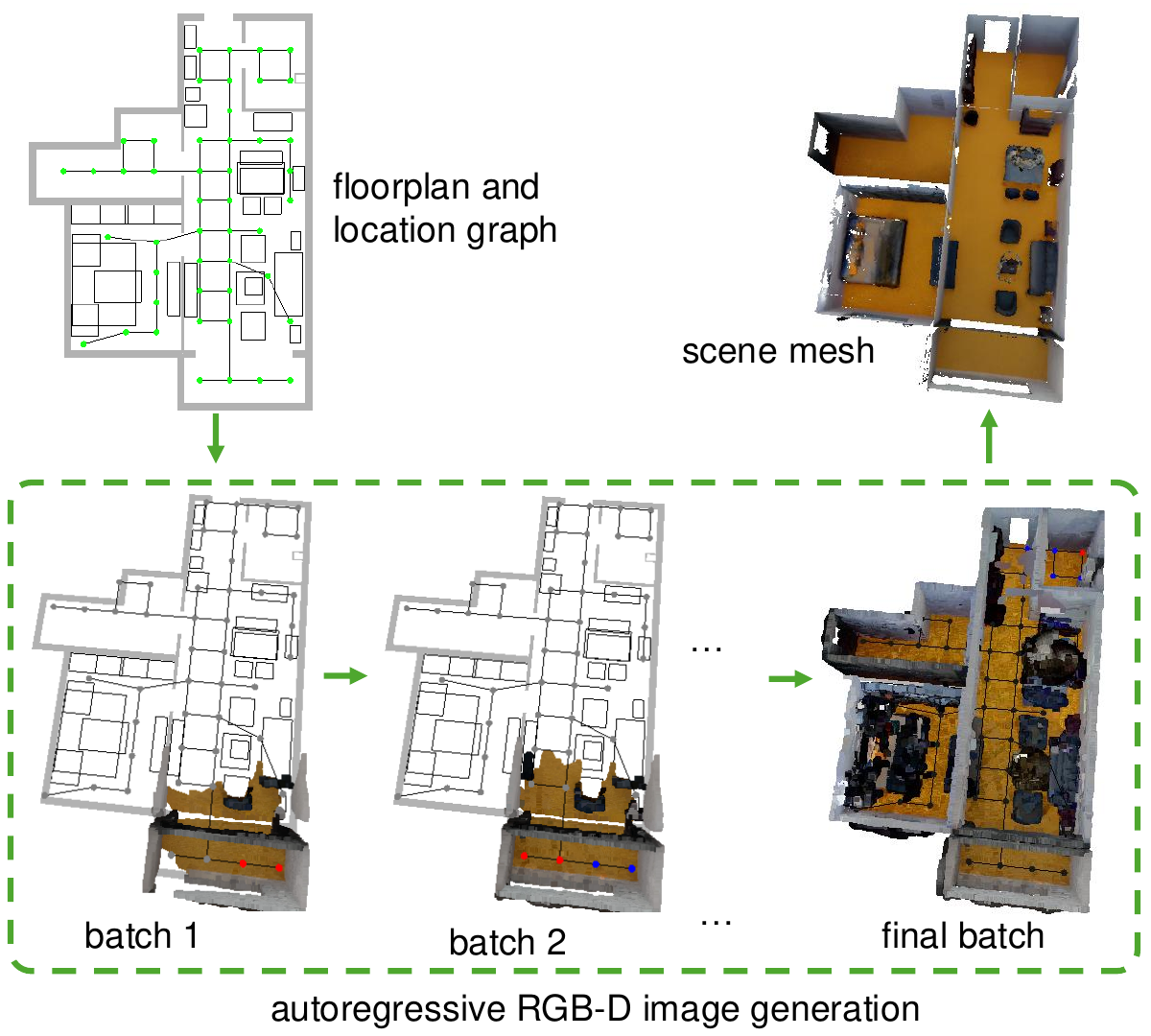}
  \vspace{-8pt}
  \caption{\textbf{Pipeline of HouseCratfter.} Given the floorplan, we sample camera locations inside the scene and construct a graph, where each node corresponds to a location (\textcolor{green}{green}). We define our generation sequence by traversing the graph. In each step, we generate RGB-D (color and depth) images at the novel view location(s) (\textcolor{red}{red}) guided by the RGB-D images already generated at the nearby visited nodes (\textcolor{black}{black}) as well as the input floorplan.
  The generated RGB-D images are converted to point cloud for visualization. After all the nodes are visited, we fuse all the generated RGB-D images into meshes.}
  \label{fig:pipeline}
  \vspace{-15pt}
\end{figure}

\section{Proposed Method: HouseCrafter}

\subsection{Overview}
Our goal is to lift a 2D floorplan to a 3D scene that we can interact with, where explicit scene representation is desired, \eg, in terms of meshes and textures.
If we had enough 3D data, training a generative model that outputs the desired 3D asset would be the most straightforward solution.
In practice, 3D data is harder to acquire and thus far more scarce than 2D imagery. 
Therefore, in this paper, we resort to generating multi-view 2D observations of the scene first and then reconstructing it in 3D.
It allows us to harness the powerful generative prior of recent advances in diffusion-based models that are trained using a large set of 2D images.

As shown in Fig.~\ref{fig:pipeline}, we sample camera locations uniformly across the free space based on the 2D floorplan and then construct a connected graph from these locations. 
The details of location sampling and graph construction can be found in  the supplementary material. 
For each location, we define a set of camera orientations to cover the surrounding. The batches in the generation sequence are decided by the order of traversing the graph (\eg, breadth-first search). To generate the first batch, thanks to the classifer-free guidance~\citep{ho2022classifier}, we only take the floorplan as condition to generate RGB-D images. When traversing the graph and encountering a node $v$ whose images have not been generated, we choose images at visited nodes within $\delta_r$ hops from $v$ as reference views, and views at unvisited nodes within $\delta_n$ hops from $v$ that as novel views (details are provided in the supplementary material).
After exhausting these locations, we use TSDF fusion ~\citep{tsdf} to reconstruct a detailed 3D vertex-colored mesh from the generated RGB-D images.

\subsection{Floorplan-guided Novel View RGB-D Image Generation}
The core of our approach is the novel view RGB-D image generation module guided by the input floorplan.
Specifically, we modify and fine-tune the UNet of the \verb|StableDiffusion v1.5|~\citep{rombach2022sd} to repurpose its powerful generation capacity obtained from training on web-scale data for our setting while keeping their VAE frozen. Specifically, given the 2D floorplan $L$, and the already generated RGB and depth images 
$\{(\imagenotation_i^r, \depthnotation_i^r)\}_{i=1}^{N_r}$ at poses $\{\campose_i^r\}_{i=1}^{N_r}$ as references, the goal of our novel view synthesis model is to generate RGB-D images $\{(\imagenotation_j^n, \depthnotation_j^n)\}_{j=1}^{N_n}$ at the novel poses $\{\campose_j^n\}_{j=1}^{N_n}$.
Here $N_r$ and $N_n$ denote the number of reference and novel images, respectively.

First, the condition information is encoded before passing to the denoising UNet. The floorplan encoding $l_j$ for each novel view is obtained from 2D floorplan $L$ and the pose $\campose_j^n$ (Sec \ref{sec:floor_condition}). The reference RGB image $\imagenotation_i^r$ is embedded to a latent feature $\imagelatentnotation_i^r$ using a lightweight image encoder~\citep{woo2023convnext} while the reference depth image $\depthnotation_i^r$
is unprojected to point cloud $\pointcloud_i^r$ (Sec \ref{sec:rgbd_condition}). From the processed condition, $\{\layoutnotation_j\}_j$, $\{\imagelatentnotation_i^r\}_i$, and $\{\pointcloud_i^r\}_i$, our modified UNet denoises the novel view latents $\{(\imagelatentnotation_j^n, \depthlatentnotation_j^n)\}_{j=1}^{N_n}$, which are then decoded to RGB-D images using the frozen VAE decoder (Sec \ref{sec:rgbd_gen}).

\begin{figure*}[!t]
  \centering
  \graphicspath{{img/}}
  \includegraphics[width = 0.9\linewidth]{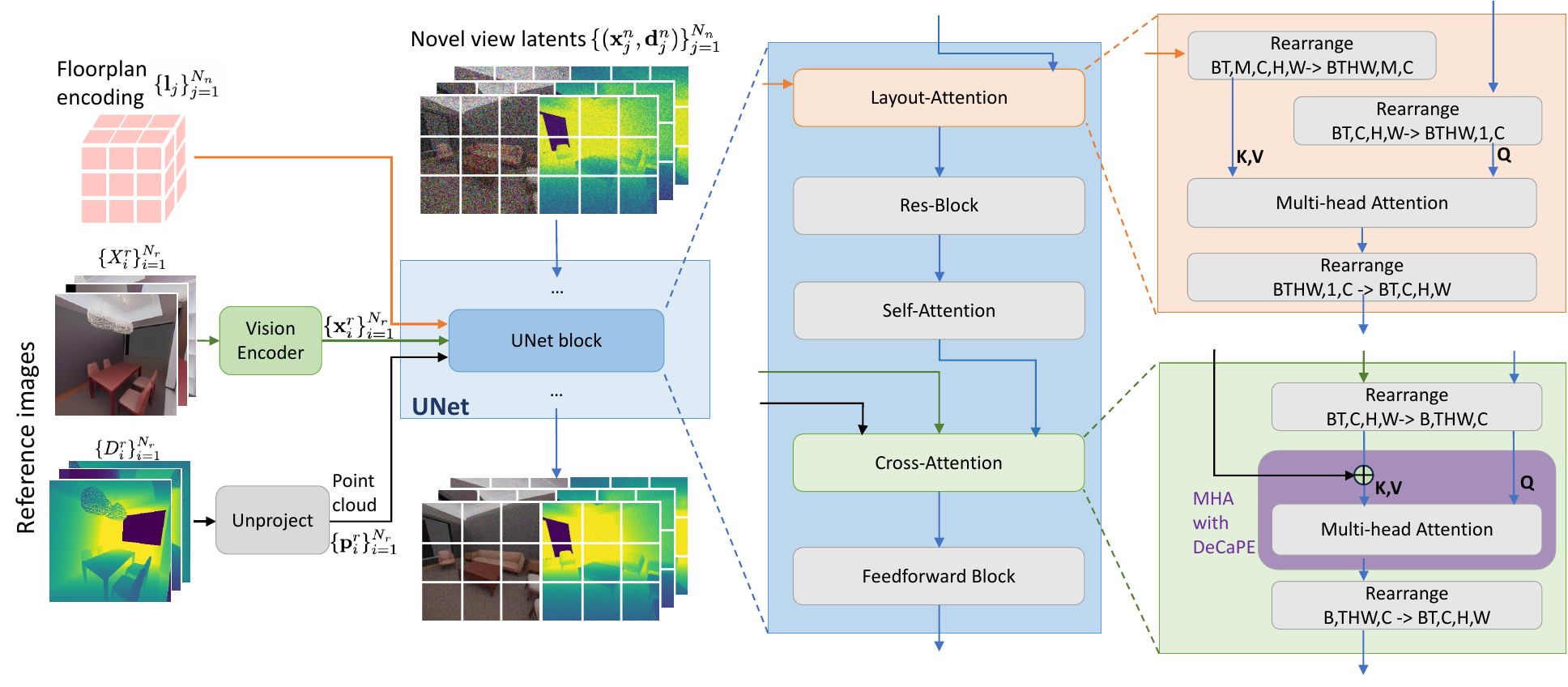}
  \vspace{-12pt}
  \caption{\textbf{Floorplan-guided novel view RGB-D generation model.} 
  First, our model simultaneously denoises the latent of RGB and depth images $\{(\imagelatentnotation_j^n, \depthlatentnotation_j^n)\}_{j=1}^{N_n}$, enabling geometry and texture consistency. Second, the introduced layout-attention block allows the novel view latent $(\imagelatentnotation_j^n, \depthlatentnotation_j^n)$ to condition on the corresponding encoded floorplan $\layoutnotation_j$. Third,  DeCaPE (the bottom of the rightmost column) is proposed to leverage the explicit geometry of the reference views to better guide the generation of RGB-D images at the novel views.
  }
  \label{fig:novel_view_model}
\end{figure*}

An illustration of the model is shown in Fig.~\ref{fig:novel_view_model}.
Our model architecture is inspired by designs of SOTA object-centric novel view synthesis models~\citep{free3d,escherNet}, but re-designed for the geometric and semantic complexity of scene-level contents.
First, we extend both the reference conditioning and image generation to the RGB-D setting instead of RGB only, as RGB-D images provide strong cues for 3D scene reconstruction.
Second, we insert a ``layout attention'' layer at the beginning of each UNet block to encourage the generated images to be faithful to the floorplan, ensuring global consistency in generating a house-scale scene.
Moreover, the cross-attention layer, which is introduced in prior works for reference-novel view attention, is updated to leverage geometry from the reference depth, leading to higher-quality image generation.

\subsubsection{Multi-view RGB-D Generation} 
\label{sec:rgbd_gen}

Given RGB and depth latents $\imagelatentnotation_j^n$ and $\depthlatentnotation_j^n$ of a novel view, 
instead of denoising them separately, we concatenate them along the channel dimension as $\diffusionlatent_j^n = [\imagelatentnotation_j^n, \depthlatentnotation_j^n]$ and denoise them jointly. In this way, the model can effectively fuse the information of RGB and depth images into a single representation to ensure the \emph{semantic} consistency between them at a single view. We double the input and output channels of the UNet to accommodate $\diffusionlatent_j^n$. When we denoise a set of latents $\{\diffusionlatent_j^n\}_{j=1}^{N_n}$ simultaneously, it ensures consistency across RGB and depth images both semantically and geometrically across different views and thus leads to higher-quality generation as shown in the experiments. 

To leverage the frozen VAE for depth images, we 
replicate the depth image to 3 channels, clip the depth to a preset of near and far planes ($[0,3]$ meters), then normalize it to the range $[-1,1]$ before passing to the VAE encoder. From the depth latent, we decode it then average over 3 channels before unnormalizing the value to the metric depth. 

\subsubsection{Floorplan Conditioning} 
\label{sec:floor_condition}

\begin{figure}
  \centering
  \includegraphics[width = 0.8\linewidth]{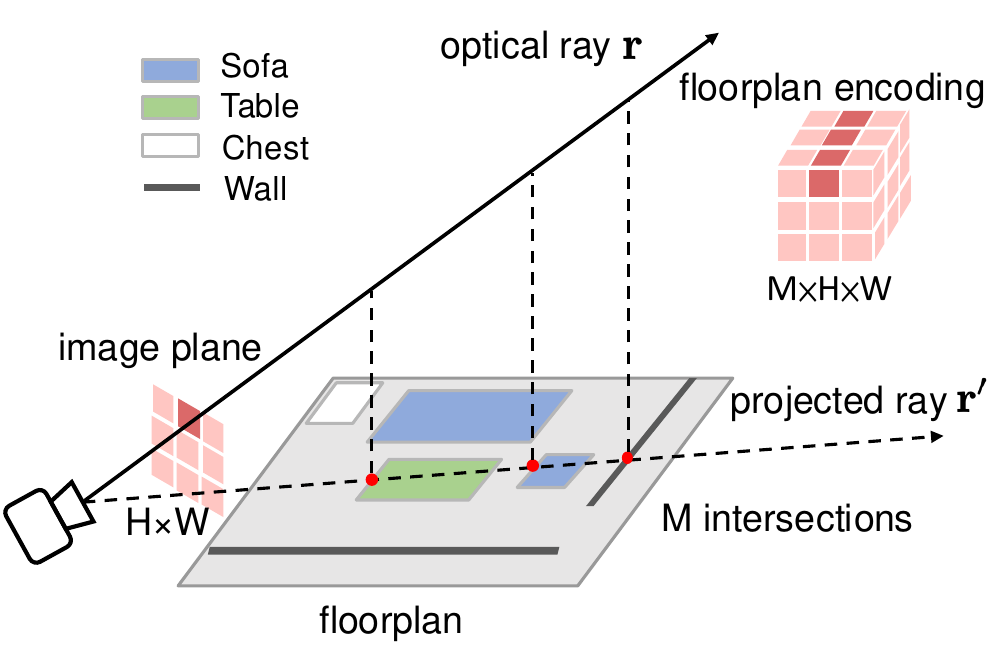}
  \caption{\textbf{Floorplan Encoding.} 
  We project the camera ray $\mathbf{r}$ to the floor plane. Along the projected ray $\mathbf{r}'$, we find the intersections with floorplan's components. For each pixel, there are at most $M$ intersections, representing potential objects that may be seen at this pixel. We embed the location and associated object class of the intersections to a latent space to obtain floorplan encoding.}
  
\label{fig:floorplan_cencoding}
\end{figure}

We use a vectorized representation $L$ for the floorplan~\citep{zheng2023layoutdiffusion}, which describes the structure and furniture arrangement of the house from a bird-eye view. $L = \{o_k\}_{k=1}^{N}$ consists of $N$ items, where each component $o_k = \{c_k,p_k\}$ is specified by its category $c_k$ and geometry information $p_i$. If the component $o_k$ represents furniture (\eg, a chair), $p_k$ defines the 2D bounding box enclosing the object. For other components, including walls, doors, and windows, it specifies the start and end point of a line segment corresponding to the them.

To use it as condition to the diffusion model, we encode the floorplan for each novel view. Fig.~\ref{fig:floorplan_cencoding} illustrates the encoding process for a novel view. 
In specific, for every pixel of the RGB-D latent $\diffusionlatent_j^n \in \mathbb{R} ^{C \times H \times W}$ at a novel view, we shoot a ray $\raynotation$ originating at the camera center of $\diffusionlatent_j^n$ and passing through the pixel center, which is then orthogonally projected to the floor plane to obtain $\raynotation'$. Along the projected ray $\raynotation'$, we take at most $M$ intersections with the 2D object bounding boxes or other floorplan components (\eg, walls). Gathering across all the pixels of $\diffusionlatent_j^n$, we get $M \times H \times W$ intersections (padding is applied for a ray with less than $M$ intersections). With each intersection point, we obtain object category and the position, resulting in  $\categorynotation_j \in \mathbb{N}^{M \times H \times W}$ for the semantic category and $\positioninfonotation_j \in \mathbb{R}^{M \times 2 \times H \times W}$ for the point position where the dimension of 2 consists of the depth along the project ray $\raynotation'$ and the \textcolor{black}{angle between ray and floor} 
Note that we exclude the intersections after the ray first hits the wall to take the occlusion into effect, and use zero-padding to ensure the same number of intersection points per ray for batching.
$\categorynotation_j$ and $\positioninfonotation_j$ are then projected to a latent $\layoutnotation_j$.
\begin{align}
    \layoutnotation_j &= \texttt{Embed}(\categorynotation_j) + \texttt{PosEnc}(\positioninfonotation_j),
\end{align}
where $\texttt{Embed()}$ map each semantic class to a latent vector and $\texttt{PosEnc()}$ is the sinusoidal position embedding. 
$\layoutnotation_j \in \mathbb{R}^{M \times C \times H \times W}$  encodes both the geometry and semantic of the floorplan.

%
%

Subsequently, the layout-attention block modulates RGB-D  latents using cross-attention between the image latents and $\layoutnotation_j$ on pixel level, where each latent feature in $\diffusionlatent_j^n$ is the query and the floorplan features along the corresponding ray are the keys and values, meaning the attention for each pixel is performed independently.
We provide more technical details in the supplementary material.


\subsubsection{Multi-view RGB-D Conditioning} 
\label{sec:rgbd_condition}
In addition to being faithful to the input floorplan, the generated RGB-D images should be consistent with the reference images as well.
Cross-attention of multi-view RGB-only images with camera poses has been investigated in prior works \citep{escherNet,miyato2023gta} for this purpose. 
However, in our case, we have the depth images from the reference views that can provide additional geometry information. Hence, we introduce Depth-enhanced Camera Positional Encoding (DeCaPE) for cross-attention between the novel views (query) and reference views (key) to improve the generation quality.

\begin{figure}[!t]
    \centering
    \includegraphics[width=0.95\linewidth]{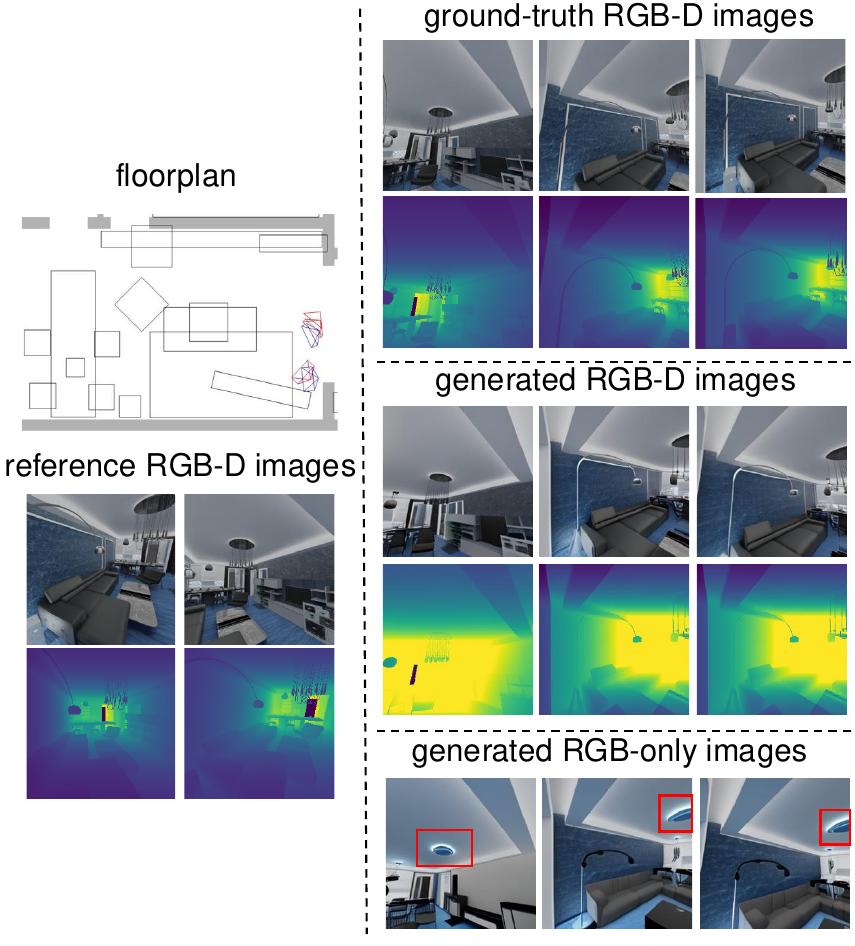}
    \vspace{-6pt}
    \caption{\textbf{Visual results of novel view RGB-D image generation.}
    \textbf{Left:} We show input floorplan and reference RGB-D images. The cameras are overlaied on top of the floorplan for better visualization, where \textcolor{blue}{blue} and \textcolor{red}{red} ones correspond to the reference and novel views, respectively.
    \textbf{Right:} We show ground-truth RGB-D images in the top, RGB-D image generation results in the middle, and RGB-only image generation results (no depth images as input either) in the bottom.
    We can see that without using depth images as conditioning (input) and output, the output has more artifacts, for instance, the inconsistent ceiling lamp w.r.t. the reference views (highlighted with red bounding boxes).
    }
    \vspace{-28pt}
    \label{fig:rgbd_gen}
\end{figure}

We first revisit Camera Positional Encoding (CaPE) proposed in \citet{escherNet} and then describe DeCaPE. 
To avoid notation clutter, let's denote $\campose_Q=\campose_j^n$ and $\campose_K=\campose_i^r$.
Further, we have $\novelviewnotation_Q$ and $\novelviewnotation_K$, which are tokens from novel view latent $\diffusionlatent_j^n$ and reference RGB latent $\imagelatentnotation_i^r$, respectively.
In CaPE, $\phi(\campose)$ is defined in analogy to camera extrinsic $\campose$ so that the high-dimensional latent vector $\novelviewnotation$ can be transformed via $\phi(\campose)$ in the similar way that point cloud coordinate is transformed via $\campose$,
\begin{equation}
    {\phi(\campose)} = 
\begin{bmatrix}
 {\campose} & 0 & \cdots & 0 \\
 0 & {\campose} & 0 & \vdots \\
 \vdots & 0 & \ddots & 0  \\
0 & \cdots & 0 &  {\campose} 
\end{bmatrix}.
\end{equation}
The similarity between $\novelviewnotation_Q$ and $\novelviewnotation_K$ is then computed as
\begin{align}
    s_{QK} &= \langle \phi(\campose_Q^{-\intercal})\novelviewnotation_Q,  \phi(\campose_K)\novelviewnotation_K \rangle 
    \\&= \novelviewnotation_Q^\intercal \phi(\campose_Q^{-1}) \phi(\campose_K)\novelviewnotation_K 
    \\&= \novelviewnotation_Q^\intercal \phi(\campose_Q^{-1}\campose_K)\novelviewnotation_K \label{eq:cape}.
\end{align}
The key property of CaPE is that $\campose_Q^{-1}\campose_K$ encodes the relative transformation of the camera poses while being invariant to the choice of the world coordinate system. 
Eq.(\ref{eq:cape}) can be interpreted as the feature of the reference view, $\novelviewnotation_K$, in its camera coordinate system is transformed to the coordinate system of the novel view, $\phi(\campose_Q^{-1}\campose_K)\novelviewnotation_K$, before taking the dot product with the query feature. Since we have the explicit 3D position of the tokens from the reference depth image, DeCaPE uses the 3D position to augment $\novelviewnotation_K$ in its camera coordinate before applying the camera transformation,
\begin{equation}
    s_{QK} = \novelviewnotation_Q^\intercal \phi(\underbrace{\campose_Q^{-1}\campose_K}_{\text{camera poses}})(\novelviewnotation_K + \underbrace{\texttt{PosEnc}(\positioninfonotation_K)}_{\text{3D position from depth}}),
    \label{eq:enhanced_cape}
\end{equation}
where $\positioninfonotation_K$ is the 3D position of $\novelviewnotation_K$  in the camera coordinate system of the key (reference view), obtained from depth image. 
While preserving the invariance to the choice of world coordinate, Eq.(\ref{eq:enhanced_cape}) enhances the similarity (attention score) computation of CaPE for the cross attention and therefore leads to better generation as we will show in the experiments.

\begin{figure*}[!htbp]
  \centering
  \includegraphics[width = 0.9\linewidth]{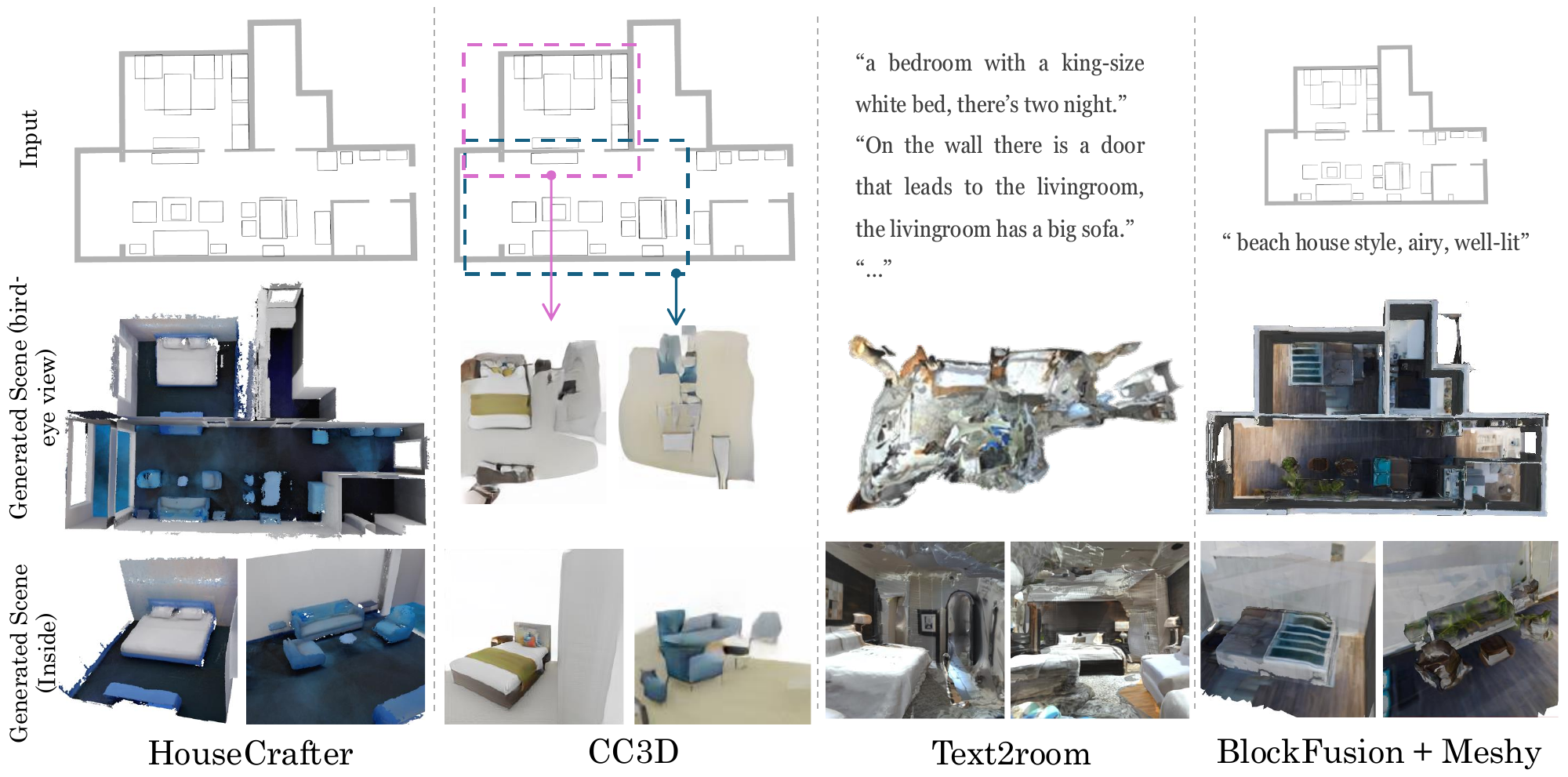}
  \vspace{-4pt}
  \caption{\textbf{Qualitative comparisons.} We show two random viewpoints for each scene as well as a top-down views. We
compare our model with CC3D~\citep{bahmani2023cc3d}, Text2Room~\citep{text2room}, and BlockFusion~\citep{blockfusion}. HouseCrafter generates results with better geometry and textures. More examples are provided in the supplementary material.
}
\vspace{-10pt}
  \label{fig:results}
\end{figure*}

%% file: sec/4_Experiments.tex
\section{Experiment}
\subsection{Experimental Setup}

\noindent\textbf{Dataset.} We conduct experiments on 3D-FRONT~\citep{front3d}, a synthetic indoor scene dataset that contains rich house-scale layouts and is populated by detailed 3D furniture models. Compared with other indoor scene datasets~\citep{dai2017scannet,chang2017matterport3d}, it allows us to render high-quality images of the scene at any selected camera poses, which is essential to training our novel view RGB-D image generation model. For each house in the dataset, we obtain the floorplan based on furniture bounding boxes and wall mesh.
Nearly 2000 houses with 2 million rendered images are used for training and 300 houses are for evaluation.

\textcolor{black}{To verify the model's ability to generalize to realworld scenes, we finetune our depth and layout condition design with stable-virtual-camera~\citep{seva} backbone using data from ARKitScene~\citep{arkitscenes}. Results are shown in Fig.\ref{fig:teaser} and supplementary material}

\noindent\textbf{Evaluation.} 
We first evaluate the quality of multi-view RGB-D image generation.
Specifically, we measure the consistency across reference-novel (\textbf{R-N}) and novel-novel (\textbf{N-N}) views. 
While the open-ended nature of the generation task makes the evaluation challenging due to the absence of ground truth information, we can measure the consistency of two views within their overlap region, which can be estimated via the depth and camera poses.
Given the estimated overlap region, we evaluate RGB consistency using PSNR and depth consistency using Absolute Mean Relative Error (AbsRel) and percentage of pixel inliers $\delta_i$ with threshold $1.25^i$. We also report Fréchet Inception 
\begin{figure}
  \centering
  \graphicspath{{img/}}
  \includegraphics[width = \linewidth]{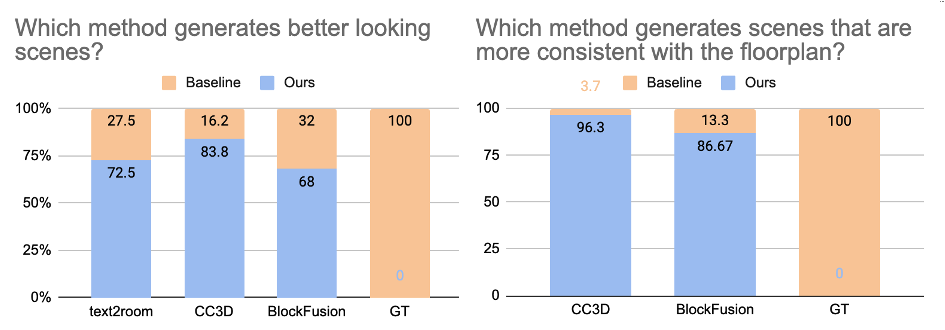}
  \caption{\textbf{User study.} ~Participants significantly favor
our method over baselines, for both overall quality and consistency w.r.t. the floorplan. 
The high preference of ground-truth (GT) results validates the trustworthiness of our user study results.
  }
  \label{fig:UserStudy}
  \vspace{-15pt}
\end{figure}
Distance (FID)~\citep{heusel2017gansfid} and Inception Score (IS)~\citep{salimans2016is} for the general visual quality. 

Next, we evaluate the faithfulness of the generated RGB-D images to the input floorplan by resorting to the state-of-the-art 3D instance segmentation method, ODIN~\citep{jain2024odin}. We extract top-down 2D boxes of the 3D segmentation results to compare with the floorplan's boxes using mAP@25 (mean Average Precision at the intersection-over-union threshold of 0.25)~\citep{coco}. While the absolute value of mAP does not directly reflect the floorplan consistency of the generated scenes due to segmentation errors, we find that mAP@25 has positive correlation with floorplan consistency, meaning better generation results leading to higher mAP. 
We also report mAP@25 of ground-truth images as a reference.

\definecolor{mygray}{gray}{0.5}
\newcommand{\gtres}[1]{\textcolor{mygray}{#1}}

\begin{table}[!t]
  \caption{\textbf{Quantitative comparisons with state-of-the-art methods for scene generation.}
  IS metrics are computed on the \emph{rendered} RGB images in the generated scenes. \textcolor{black}{3D geometry quality is measured by the number of connected components, which is a metric to evaluate the severity of dangling artifacts} We also show the results of using ground-truth images from 3D-FRONT as references.
  }
  \label{depth2d}
  \centering
  \small
  \setlength{\tabcolsep}{1.5pt}

  \begin{tabular}{ccccl}
  \toprule
  \label{tab:baseline}
  \multirow{2}{*}{Method} & Visual Quality & Floorplan Consistency  &3D quality\\ 
    \cmidrule{2-4}
    & IS $\uparrow$ & mAP@25$\uparrow$  & artifacts$\downarrow$\\ 
    \midrule
    Text2Room & \textbf{5.35}& -  &302.5\\
     CC3D & 4.02 &  25.60  &-\\
     BlockFusion & 5.01 &  0.81  &118.64\\
     \textbf{HouseCrafter} & 4.37 & \textbf{45.77}    &178.0\\
     \midrule
     \gtres{GT, 3D-FRONT} & \gtres{4.50} & \gtres{54.51}  &-\\ 
    \bottomrule
  \end{tabular}
  \vspace{-15pt}
\end{table}

Finally, we evaluate the quality of generated 3D scenes.
To this end, we first conduct a user study, involving 12 participants, to compare our approach with baseline methods in terms of perceptual quality and coherence to the given floorplan. For each baseline, 8 pairs of meshes (our vs. baseline) are shown to the participants. We also add 3 pairs with ground-truth meshes, as a means to justify the trustworthiness of the user study results, resulting in a total of 228 data points. 
In addition, we report IS calculated from RGB images \emph{rendered} at random camera poses for each generated scene. 
We provide more details about evaluation in the supplementary material.

\begin{table*}[t]

  \caption{\textbf{Ablation studies of different design choices.}
  We evaluate novel view RGB-D image generation and the consistency of the generated scene to the input floorplan (Floor. Const.). The best results are highlighted with \textbf{bold} and the second best with \underline{underline}.}
  \label{tab:ablation_rgbd_nvs}
  \centering
  \small
  \setlength{\tabcolsep}{6pt}
\begin{tabular}{
    c      
    c      
    c      
    c      
    c      
    r      
    r      
    r      
    r      
    r      
    r      
    r      
    r      
    r      
  }
    \toprule
    \multirow{3}{*}{Variant} & 
    \multirow{3}{*}{\begin{tabular}{@{}c@{}}Output \\ Depth\end{tabular}} &
    \multirow{3}{*}{\begin{tabular}{@{}c@{}}Input \\ Depth\end{tabular}} &
    \multirow{3}{*}{\begin{tabular}{@{}c@{}}Floorplan \\ Cond.\end{tabular}} & 
    \multirow{3}{*}{\begin{tabular}{@{}c@{}}w/DeCaPE\end{tabular}} &
    \multicolumn{4}{c}{RGB Metrics} & \multicolumn{4}{c}{Depth Metrics} & Floor. Const. \\ 
    \cmidrule(lr){6-9} \cmidrule(lr){10-13} \cmidrule{14-14}
     &   &  &  &  &   \multirow{2}{*}{FID $\downarrow$} & \multirow{2}{*}{IS $\uparrow$} & \multicolumn{2}{c}{PSNR $\uparrow$} & \multicolumn{2}{c}{AbsRel $\downarrow$} & \multicolumn{2}{c}{$\delta_{0.5}$$\uparrow$} & \multirow{2}{*}{mAP@25$\uparrow$} \\
    \cmidrule(lr){8-9}  \cmidrule(lr){10-11}  \cmidrule(lr){12-13}
     &  &  & & & && R-N & N-N & R-N & N-N & R-N & N-N & \\
    \midrule
    \ding{172}& \xmark  & \xmark & \xmark  & \xmark & 49.35 & 5.00 & -&  - & - & - & - & -  & - \\
    \ding{173} & \cmark & \xmark   & \xmark    & \xmark &33.39&\textbf{5.23}& 20.99 & 22.60 & 23.56  & 11.48 & 79.14 & 88.79 & - \\
    \ding{174} &\cmark & \cmark & \xmark  & \xmark &35.77&\underline{5.16}& 20.91 & 21.98 & 22.28 & 12.05 & 81.78 & 88.23 & - \\
    \ding{175} &\cmark& \xmark   & \cmark  & \xmark &\textbf{15.64}& 4.70 & \textbf{25.36} & \textbf{24.79} & 7.65 & \underline{7.85} & \underline{90.44} & \underline{91.77} & 48.46  \\
    \ding{176} &\cmark& \cmark & \cmark  & \xmark &\underline{16.70}& 4.70& 24.47 & 24.68 & \underline{7.73} & 8.20 & - & - & 49.74 \\
    \ding{177} &\cmark& \cmark & \cmark  & \cmark &17.75& 4.74& \underline{25.31} & \underline{24.69} & \textbf{6.79} & \textbf{7.37} & \textbf{92.20} & \textbf{92.65} & \textbf{52.26} \\
    \bottomrule
\end{tabular}

  \vspace{-12pt}
\end{table*}

\subsection{Comparisons with state of the art} 

\noindent\textbf{Baselines.}
To the best of our knowledge, there are no direct methods that generate 3D houses from 2D floorplans. The closest works to ours are CC3D~\citep{bahmani2023cc3d} and BlockFusion~\citep{blockfusion}, which produce a scene from 2D layout. CC3D represents the scene as a feature volume that can be rendered with a neural renderer to obtain RGB and depth images. 
Since BlockFusion only generates a scene mesh, an text-to-texture method, Meshy\footnote{https://www.meshy.ai/}, is used to texturize the generated mesh. We also compare against Text2Room~\citep{text2room}, which generates an indoor scene from a series of text prompts. 
As Text2Room~\citep{text2room} does not receive any floorplan guidance, we only compare to it in terms of visual quality. 

\noindent\textbf{Results.}
We provide qualitative results in Fig.~\ref{fig:results} and quantitative comparisons in Fig.~\ref{fig:UserStudy} and Table~\ref{tab:baseline}.
We refer readers to the supplementary material for more visual comparisons.
For the visual quality (IS metric), both human (Fig.~\ref{fig:UserStudy}) and automated (Table~\ref{tab:baseline}) evaluations show that our method performs better than CC3D. 
However, Text2Room and BlockFusion report better IS metrics than our method, while the users significantly prefer our results.
We believe their higher IS metrics are due to the more diverse scenes generated by using models trained using the real-world data  (\eg, the pre-trained text-to-image model~\citep{rombach2022sd} in TextRoom), which have similar distribution to the data used to train the model for IS evaluation.
This is evidenced by the lower IS metric when using the ground-truth images from 3D-FRONT (``GT, 3D-FRONT'' in Table~\ref{tab:baseline}). 

In terms of the consistency w.r.t. the input floorplan, as we explained earlier, although the mAP@25 metric is not perfect due to its dependence on the 3D instance segmentation model, it correlates with the floorplan consistency positively.
As seen in the last row in Table~\ref{tab:baseline}, using GT images leads to the best mAP@25 score.
Compared with CC3D and BlockFusion, our method performs better reflected in the higher mAP@25.
The effectiveness of our approach is further validated by the user study results reported in Fig.~\ref{fig:UserStudy}.

\subsection{Ablation Studies}


We conduct ablations on various design choices of our RGB-D image generation model using 300 test houses from 3D-FRONT~\citep{front3d}.
For measuring RGB-D generation quality, camera poses are sampled in groups of 6, with 3 as reference and 3 as novel views. We measure consistency between reference-novel (R-N) and novel-novel (N-N) pairs. For floorplan consistency evaluation, we use images generated in the full autoregressive pipeline. 


\noindent\textbf{Generating depth images improves RGB image generation.} Variant pair (\ding{172}, \ding{173}) in Table~\ref{tab:ablation_rgbd_nvs} demonstrates that generating depth alongside RGB improves FID and IS for RGB images, 
validating our motivation of generating RGB-D images together.

\noindent\textbf{Depth conditioning enhances geometry consistency.} 
As seen in (\ding{173},\ding{174}) and (\ding{175},\ding{176}), using input depth enhances R-N and N-N depth consistency, with a larger gain for R-N, although RGB consistency is less affected Geometry improvements also translate to better floorplan consistency.

\noindent\textbf{DeCaPE further enhances geometry.} (\ding{176}, \ding{177}) show our DeCaPE encoding yields higher depth and floorplan consistency than CaPE, which further helps generating floor plan with better geometric quality.

\noindent\textbf{Floorplan guidance is critical for both color and geometry quality.} 
(\ding{173}, \ding{175}) and (\ding{174}, \ding{176}) show that floorplan conditioning strongly improves all metrics, especially for depth. Despite limited 3D cues, our 2D floorplan input still enables high-quality 3D scene generation.


%% file: sec/5_Conclusion.tex
\section{Conclusion}




In this work, we presented HouseCrafter, a model that can transform 2D floorplans into detailed 3D scenes. We generate dense RGB-D images autoregressively and fuse them into a 3D mesh. Our key innovation is an image-based diffusion model that produces multiview-consistent RGB-D images guided by floorplan and reference RGB-D images. This capability enables the generation of house-scale 3D scenes with high-quality geometry and texture, empowering downstream applications in other domains.

\newpage

%% file: sec/Appendix.tex
\newpage
\clearpage
\setcounter{page}{1}

\twocolumn[{%
\renewcommand\twocolumn[1][]{#1}%
\maketitlesupplementary
\centering
    \graphicspath{{img/}}
  \includegraphics[width = 0.95\linewidth]{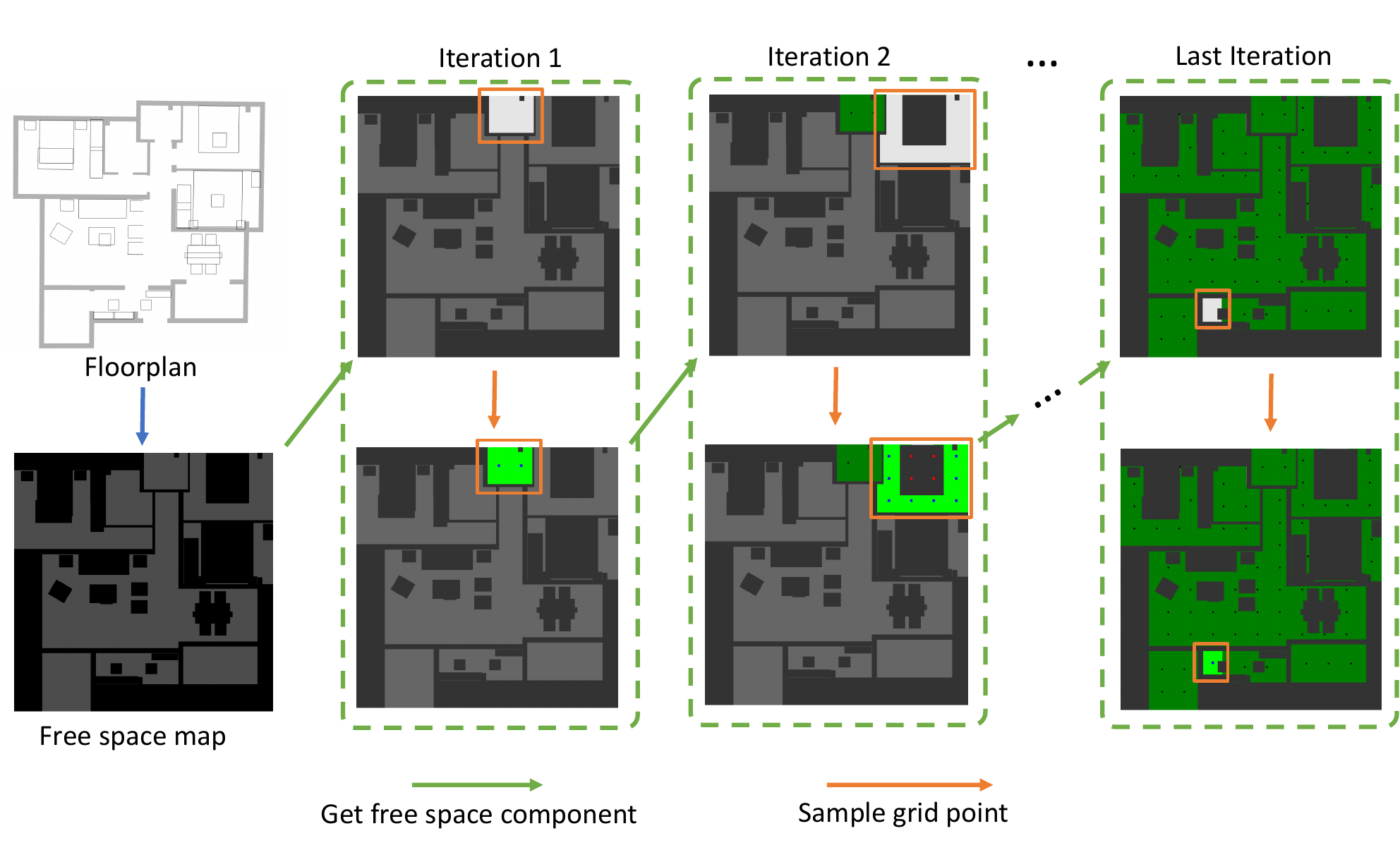}
  \captionof{figure}{\textbf{Camera location sampling.} From the floorplan, we first obtain a binary free space map (black means occupied) and then iterate over the free space to sample camera locations.
  In each iteration, we select a free space component (highlighted in white) then sample grid points within the component's bounding box. The invalid points (occupied, marked with \textcolor{red}{red}) are discarded and the rest valid points (unoccupied, marked with \textcolor{green}{green}) are stored as possible camera locations.  
  The loop terminates when the all the free space is processed or the remaining area is smaller than a threshold.
  }
  \label{fig:point_sampling}
    \vspace{10pt}
}]


\begin{figure*}[!htbp]
  \centering
  \graphicspath{{img/}}
  \includegraphics[width = 0.9\linewidth]{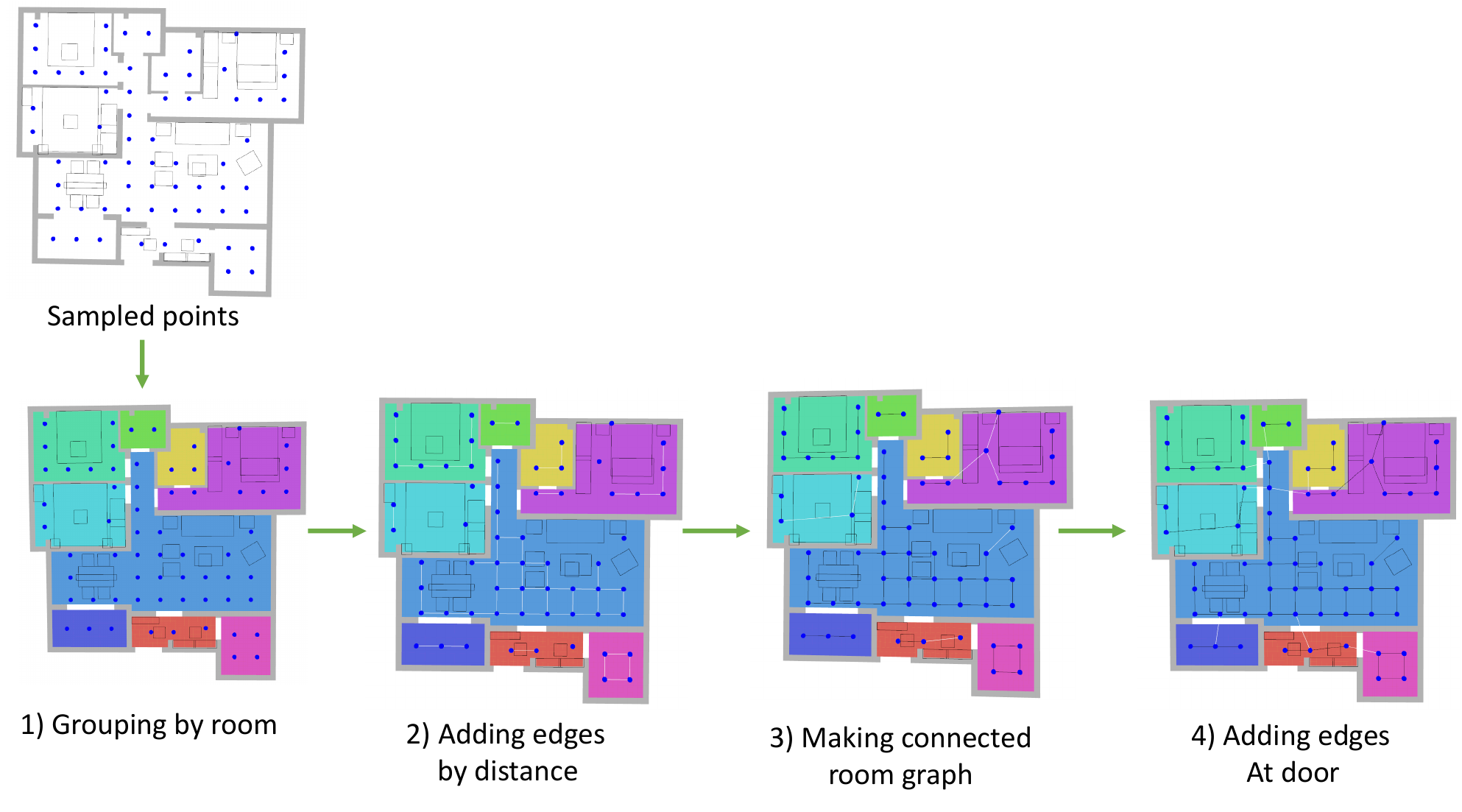}
  \caption{\textbf{Location graph construction.} From left to right: 
  (1) Given the sampled locations, we first group the locations by room types (\eg, kitchen, living room). 
  Next, we construct a subgraph in each room in two steps: (2) adding edges between two nodes if their distances are smaller than a threshold and  
  (3) connecting the rest isolated locations to its neighboring nodes in a partial subgraph. 
  (4) In the last step, we use the door locations to connect the room subgraphs to form a complete graph for the entire scene. 
  Specifically, for each door, we add an edge between the nearest locations across two adjacent rooms. By creating graphs at the room scale then connecting them using the door location, we avoid making undesirable edges where two locations are close but do not have overlap due to the wall. New edges of each step (2,3,4) are highlighted in white.}
  \label{fig:graph_construction}
\end{figure*}

\textbf{This supplementary document is structured as follows:}

    


\begin{itemize}
    \item \hyperref[sec:point_and_graph]{Camera Location Sampling and Graph Construction}

    \item \hyperref[sec:gen_sequence]{Autoregressive RGB-D Image Generation via Graph Traversal}
    
    \item \hyperref[sec_t:floorplan_cond]{Details of Floorplan Conditioning}

    \item \hyperref[sec:details_eval]{Details of Evaluation}
    \begin{itemize}
        \item \hyperref[subsec:consistency_eval]{RGB-D Image Consistency Evaluation}

        \item \hyperref[subsec:floorplan_eval]{Floorplan Consistency Evaluation}

        \item \hyperref[subsec:user_study]{User Study}
    \end{itemize}
    
    \item \hyperref[sec:floor_baseline]{Baseline for Floorplan Encoding}

    \item \hyperref[sec:implementation_details]{Implementation Details}

    \item \hyperref[sec:run_time_comp]{Running Time Comparison with Other Methods}

    \item \hyperref[sec:monocular_depth]{Ablation on simultaneous RGB and Depth Image Generation}

    \item \hyperref[sec:floor_baseline]{Ablation on Floorplan Encoding}

    \item \hyperref[sec:limitation]{Limitations and Future Directions}

    \item \hyperref[sec:additional_vis_comp]{Additional Visual Comparisons}
\end{itemize}

\section{Camera Location Sampling and Graph Construction}
\label{sec:point_and_graph}
The camera location sampling procedure is illustrated in Fig.~\ref{fig:point_sampling}, where we first obtain a binary free space map from the input floorplan.
An iterative procedure is then applied over the free unoccupied space to sample camera locations until all the free space is processed.
Based on the sampled camera locations, the graph construction is illustrated  in Fig.~\ref{fig:graph_construction}, where we first construct subgraphs within each room separately.
They are finally connected to form a complete graph of the entire scene.



\section{Autoregressive RGB-D Image Generation via Graph Traversal}
\label{sec:gen_sequence}
Given the location graph, the reference and novel views are selected by traversing the graph. The procedure is described in Alg.~\ref{al:gen}. To control the number of views in each generation step, we use two hyper parameters $\delta_r$ and $\delta_n$, which are the hop distance thresholds with respect to the current nodes for the reference and novel views, respectively. When visiting a location $v$ whose images have not been generated, we choose locations that are within $\delta_r$ hops from $v$ and have images already generated as reference views.
The novel views include $v$ and those that have not been generated and within $\delta_n$ hops from $v$. 

\renewcommand{\algorithmicrequire}{\textbf{Input:}}
\begin{algorithm*}[h]
\caption{Autoregressive RGB-D image generation via graph traversal}\label{alg:cap}
\begin{algorithmic}
\Require \\
$G(V,E)$: location graph \\
$\delta_n$: Hop distance threshold for novel views \\
$\delta_r$: Hop distance threshold for reference views \\
$L$: Floorplan
\\\hrulefill
\State $X \gets \emptyset$  \Comment{Initialize the set of visited locations.}

\For{$v$ in $BFS(G)$}  \Comment{traverse graph via breadth-first search.}
\If{$v \notin X$}
    \State $X_r \gets X \cap N(v,G,\delta_r)$ \Comment{Get reference locations. $N(v,G,d)$: nodes within $d$ hop from $v$}
    \State $X_n \gets \{v\} \cup N(v,G,\delta_n) \backslash X$ \Comment{Get novel locations.} 
    \If{$X_n \neq \emptyset$}
        
        \State $Generate\_RGBD\_Images(X_r, X_n, L)$ \Comment{Generate views at locations. For the first loop, $X_r$ can be empty.}
        \State $X \gets X \cup X_n$
    \EndIf

\EndIf
\EndFor
\end{algorithmic}
\label{al:gen}
\end{algorithm*}

\section{Details of Floorplan Conditioning}
\label{sec_t:floorplan_cond}
For a novel view with the latent feature $\diffusionlatent_j^n \in \mathbb{R} ^{C \times H \times W}$ (where $C$ is the feature dimension and $H \times W$ the spatial dimensions), we obtain the floorplan information $\layoutnotation_j \in \mathbb{R}^{M \times C \times H \times W}$  at the (latent) pixel-level by casting rays through the pixels and encoding semantic and geometric information at every intersection point between the projected ray and floorplan components.

Subsequently, we use cross-attention at the ray-level where each pixel feature the in $\diffusionlatent_j^n$ is the query and the floorplan features along the ray are the keys and values, meaning the attention for each ray is performed independently. To illustrate the operation, we add the batch dimension $B$ and use \verb|einops|~\citep{rogozhnikov2022einops} notation:
\begin{align*}
    \diffusionlatent_j^n &\leftarrow \textrm{rearrange}(\diffusionlatent_j^n, \textrm{ B C H W }\rightarrow \textrm{(B H W) 1 C})\\
    \layoutnotation_j &\leftarrow \textrm{rearrange}(\layoutnotation_j, \textrm{B N C H W} \rightarrow \textrm{(B H W) N C})\\
   \diffusionlatent_j^n &\leftarrow \textrm{MHA}(q=\diffusionlatent_j^n, k=\layoutnotation_j, v=\layoutnotation_j) \\
    \diffusionlatent_j^n &\leftarrow \textrm{rearrange}(\diffusionlatent_j^n, \textrm{(B H W) 1 C} \rightarrow \textrm{ B C H W }),
\end{align*}where \verb|MHA()| is the multihead attention.
The floorplan information is incorporated in the first block of each feature level in the UNet blocks of the image diffusion model. Since each level operates at a different resolution, this process effectively injects the encoded floorplan at multiple scales.
\section{Details of Evaluation}
\label{sec:details_eval}

\subsection{RGB-D Image Consistency Evaluation}
\label{subsec:consistency_eval}
In this section, we describe the overlap region estimation for a pair of posed RGB-D images. Then we provide the details of the evaluation metrics.

\textbf{Overlap Region Estimation.}
Given a pair of RGB-D images, $(I_1, D_1)$ and $(I_2, D_2)$, we warp images $I_1, D_1$ of the first view to the second view according to the relative camera pose between them, obtaining $I_{1 \rightarrow 2}, D_{1 \rightarrow 2}$. 
The overlap/correspondence region $\mathcal{M}$ is defined as follows such that the warped depth $D_{1 \rightarrow 2}$ match perfectly with $D_2$,
\begin{equation}
    \mathcal{M} \coloneq \mathbbm{1}(D_{1 \rightarrow 2}=D_2), 
\end{equation}
where $\mathbbm{1}()$ is the indicator function. To account for the potential inconsistency of the generated images, we introduce a tolerance threshold $\tau$ to estimate the overlap region,
\begin{equation}
    \mathcal{\hat{M}} \coloneq \mathbbm{1}(|D_{1 \rightarrow 2} - D_2| < \tau).
\end{equation}
Given the estimated overlap region $\mathcal{\hat{M}}$, the consistency metrics are computed for depth image pair $(D_{1 \rightarrow 2}, D_2)$ and the RGB image pair $(I_{1 \rightarrow 2}, I_2)$.

\textbf{RGB Consistency Metrics.}
Given the RGB image pair $(I_{1 \rightarrow 2}, I_2)$ and the overlap region $\mathcal{\hat{M}}$, we compute the peak signal-to-noise ratio PSNR for color consistency,
\begin{equation}
    PSNR \coloneq 20 \cdot \log_{10} (255) - 10 \cdot \log_{10} (MSE),
\end{equation}
where
\begin{equation}
    MSE \coloneq \frac{1}{\sum_{k} \mathcal{\hat{M}}(k)} \sum_{k} \mathcal{\hat{M}}(k) \cdot [I_{1 \rightarrow 2}(k) - I_2(k)]^2.
\end{equation}
Here $k$ is pixel index.
Note that we omit averaging over the color channel to simplify the notation.

\textbf{Depth Consistency Metrics.}
Given the depth image pair $(D_{1 \rightarrow 2}, D_2)$ and the overlap region $\mathcal{\hat{M}}$, we compute Absolute Mean Relative Error (\emph{AbsRel}) and percentage of pixel inliers $\delta_i$ for depth consistency. \emph{AbsRel} is calculated as:
\begin{equation}
    AbsRel \coloneq 
    \\\frac{1}{\sum_{k} \mathcal{\hat{M}}(k)} \sum_{k} \mathcal{\hat{M}}(k) \frac{|D_{1 \rightarrow 2}(k) - D_2(k)|}{D_2(k)}.
\end{equation}
The percentage of pixel inliers $\delta_i$ is calculated as:
\begin{align}
    \delta_i \coloneq 
    & \frac{1}{\sum_{k} \mathcal{\hat{M}}(k)} \sum_{k} \mathcal{\hat{M}}(k) \cdot \\
    & \mathbbm{1}\left(\max \left(\frac{D_{1 \rightarrow 2}(k)}{D_2(k)}, \frac{D_2(k)}{D_{1 \rightarrow 2}(k)}  \right) < 1.25^i \right).
\end{align}
We choose $i=0.5$ to have a tight threshold.

\subsection{Floorplan Consistency Evaluation}
\label{subsec:floorplan_eval}
The floorplan evaluation protocol is the ``inverse" of HouseCrater where we predict the top-down 2D bounding boxes of objects in the generated scene and compare their consistency with the provided floorplan. 
Specifically, the detected 2D bounding boxes are compared with 2D boxes from the given floorplan using mean Average Precision at the intersection-over-union threshold of 0.25 (mAP@25). 
Here, we use ODIN~\citep{jain2024odin}, a 3D instance segmentation method that takes multi-view posed RGB-D images as input and predicts instance segmentation of the point cloud accumulated from input images. Then, top-down 2D boxes are extracted from the segmented instances. As a scene may have up to 2000 images based on its size, we cannot pass all the images to ODIN at once. Instead, these images are grouped by room types and we do segmentation per room. This strategy does not affect the evaluation results since an object in the scene do not span in more than one room. We finetune ODIN on 3D-Front dataset to make the segmentation results more reliable since both HouseCrafter and CC3D~\cite{bahmani2023cc3d} are trained on this dataset.

\subsection{User Study}
\label{subsec:user_study}

\begin{figure}
  \centering
  \graphicspath{{img/}}
  \includegraphics[width = \linewidth]{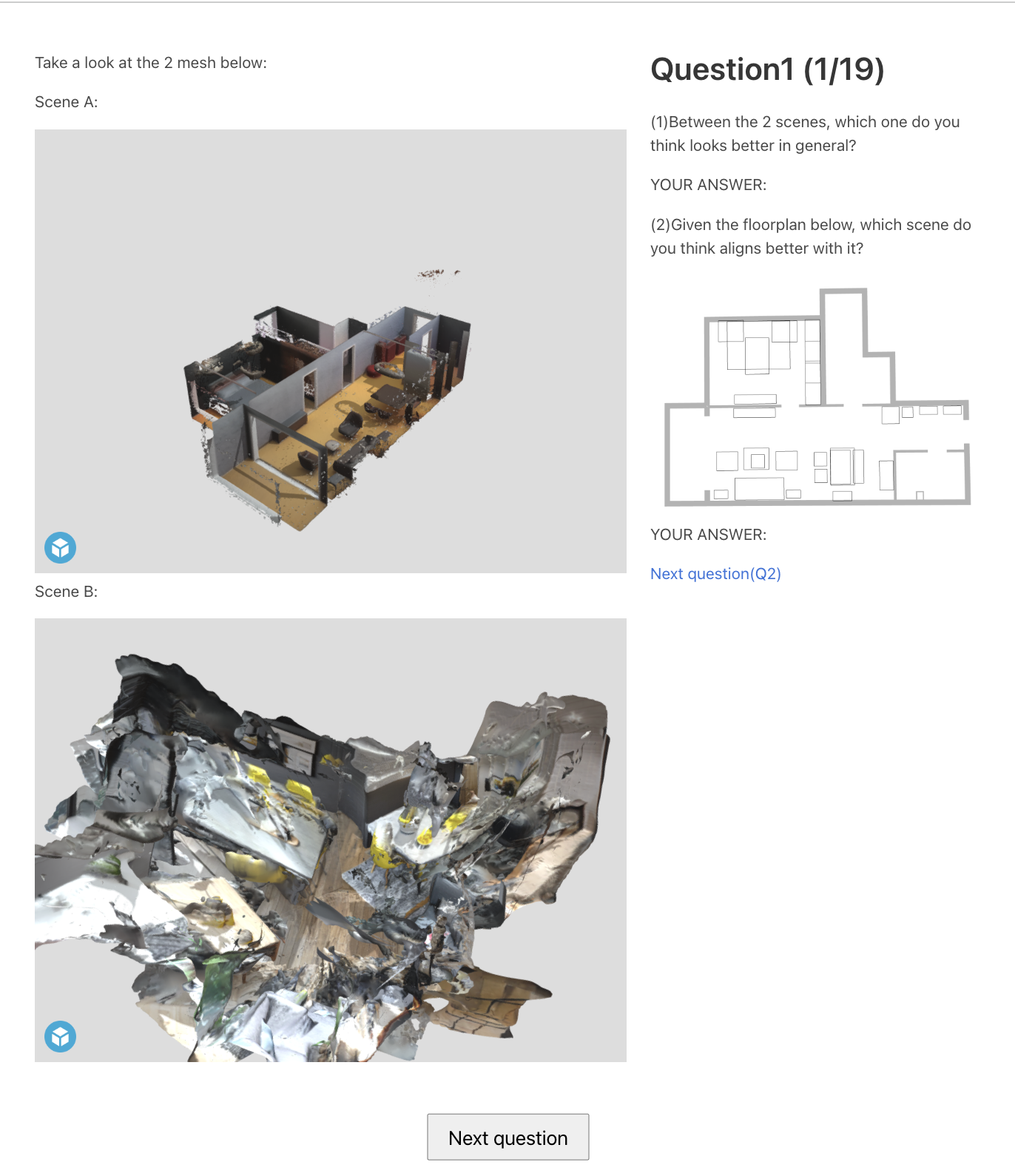}
  \caption{\textbf{User Study Interface.} We show users 2 meshes at a time, one is produced by our model and the other is produced by a baseline method. We then ask users to choose one mesh that appears "better looking in general", and one mesh that appears "align better" with the given floorplan.}
  \label{fig:interface}
\end{figure}

We conduct a user study to evaluate the results produced by Text2Room, CC3D, BlockFusion, and our method. In the study, we ask 12 participants to rate the results in a pair-wise manner. Specifically, we present the participants with two meshes at a time and ask them to choose: i) the one that appears more visually appealing; and ii) the one that is more coherent with the provided floorplan. The interface is shown in Fig.~\ref{fig:interface}. We compare the entire house we generated to the results by BlockFusion. For text2room, since it does not take floorplan as a form of guidance, we do not report participants' answers to the second question if one of the meshes is produced by it.

However, we still ask the question to prevent unconscious bias. Given that CC3D generates results at the room level rather than for entire houses, we clip our results and floorplan to the specific room CC3D produces when making comparisons.

\section{Implementation Details}
\label{sec:implementation_details}

\begin{table}[t]
    \setlength{\tabcolsep}{2pt}
  \caption{\textbf{Effectiveness of fine-tuning the novel view RGB-D generation model on noisy data for autoregressive inference.}
  }
  \label{noisy_finetune}
  \centering
  \small
  \begin{tabular}{ccclllccl}
    \toprule
    \multirow{2}{*}{Variant} & \multirow{2}{*}{Autoreg.} & FT &\multicolumn{3}{c}{RGB Metrics}& \multicolumn{2}{c}{Depth Metrics}\\
    \cmidrule(lr){4-6} \cmidrule(lr){7-8}
     & & w/ noise& FID$\downarrow$ & IS$\uparrow$ & PSNR$\uparrow$ & AbsRel$\downarrow$ & $\delta_{0.5}$$\uparrow$ \\
    \cmidrule(lr){1-1} \cmidrule(lr){2-2} \cmidrule(lr){3-3} 
    \cmidrule(lr){4-6} \cmidrule(lr){7-8}
    \ding{176} & \xmark & \xmark &16.70 & 4.74 & 25.0 & 7.06 & 92.43\\
    \ding{176} & \cmark & \xmark & 34.98 & 4.24 & 19.64 & 12.74 & 86.35\\
    \ding{177}& \cmark & \cmark & 22.30 & 4.37 & 21.76 & 11.09  & 88.13\\
    \bottomrule
  \end{tabular}
\end{table}

We initialize our novel view RGB-D image generation model from \verb|StableDiffusion v1.5|~\citep{rombach2022sd}. For the first layer of the UNet, we duplicate the pre-trained weights and divide the weights by two to accommodate the depth's latent and to reduce the change of the output scale. For the last layer of the UNet, we only duplicate the pre-trained weights for the RGB and depth output, respctively. The model is trained for $15,000$ iterations in 2 days with an effective batch size of 256 ($4$ samples per GPU $\times 8$ GPUs $\times 8$ gradient accumulation steps). Each data sample contains 3 reference views and 3 novel views with the resolution of 256. We use Adam optimizer with a learning rate of $10^{-4}$. All training is conducted on a machine with $8$ A6000 48GB GPUs.

However, during inference, instead of using ground-truth reference images, we condition the model on its previously generated outputs in an autoregressive manner. 
Since novel view images are inherently imperfect, small discrepancies in generated images can compound over multiple iterations, leading to error accumulation and progressive degradation in image quality.

To mitigate this issue, we introduce a noisy reference finetuning strategy to bridge the domain gap between training and inference. 
To this end, we construct a noisy dataset by recording the model’s generated outputs from the previous epoch. During finetuning, a subset of reference images is sampled from this noisy dataset instead of the ground-truth images.
Note the model is still trained to produce outputs that match the ground-truth novel view RGB-D images. This approach improves robustness to accumulated errors and enhances the model’s performance in long-horizon autoregressive generation.

Table~\ref{noisy_finetune} demonstrates that, compared to single-batch inference, images generated in an autoregressive manner exhibit decreased consistency and visual quality due to accumulated errors over iterations . Finetuning the model on noisy reference inputs enhances its robustness, leading to improved stability and coherence in long-horizon generation.

\section{\textcolor{black}{Running Time Comparison with Other Methods}}
\label{sec:run_time_comp}
\textcolor{black}{We measure the total time to generate a scene on an A6000 GPU. We also provide the average number of images/blocks per scene. Note that while Text2Room, BlockFusion, and HouseCrafter produce meshes as final output, CC3D generates volumetric latent as scene representation, and requires neural rendering to get any view. Hence we follow their codebase to generate a room then render 40 images and report the total time of generation and rendering. As shown in Table~\ref{tab:timebenchmark}, CC3D is the fastest method.
Among the rest, which are diffusion-based methods, our model has the smallest running time. 
}

\begin{table}[t]
  \caption{\textcolor{black}{\textbf{Running time comparison.}
  * denotes the number of blocks}}
  \centering
  \small
  \begin{tabular}{lcc}
  \toprule
  \label{tab:timebenchmark}
    Method & \#Images/Blocks & Total time (min)\\ 
    \midrule
    Text2Room & $217_{\pm 5}$& $50_{\pm 1}$  \\
     CC3D & $40$ & $<1$ \\
     BlockFusion & *$23_{\pm 10}$ & $30_{\pm 12}$\\
     HouseCrafter (Ours) & $1000_{\pm 400}$ & $24_{\pm 10}$   \\
    \bottomrule
  \end{tabular}
  
\end{table}

\begin{figure*}[t]
  \centering
  \graphicspath{{img/}}
  \includegraphics[width = 0.8\linewidth]{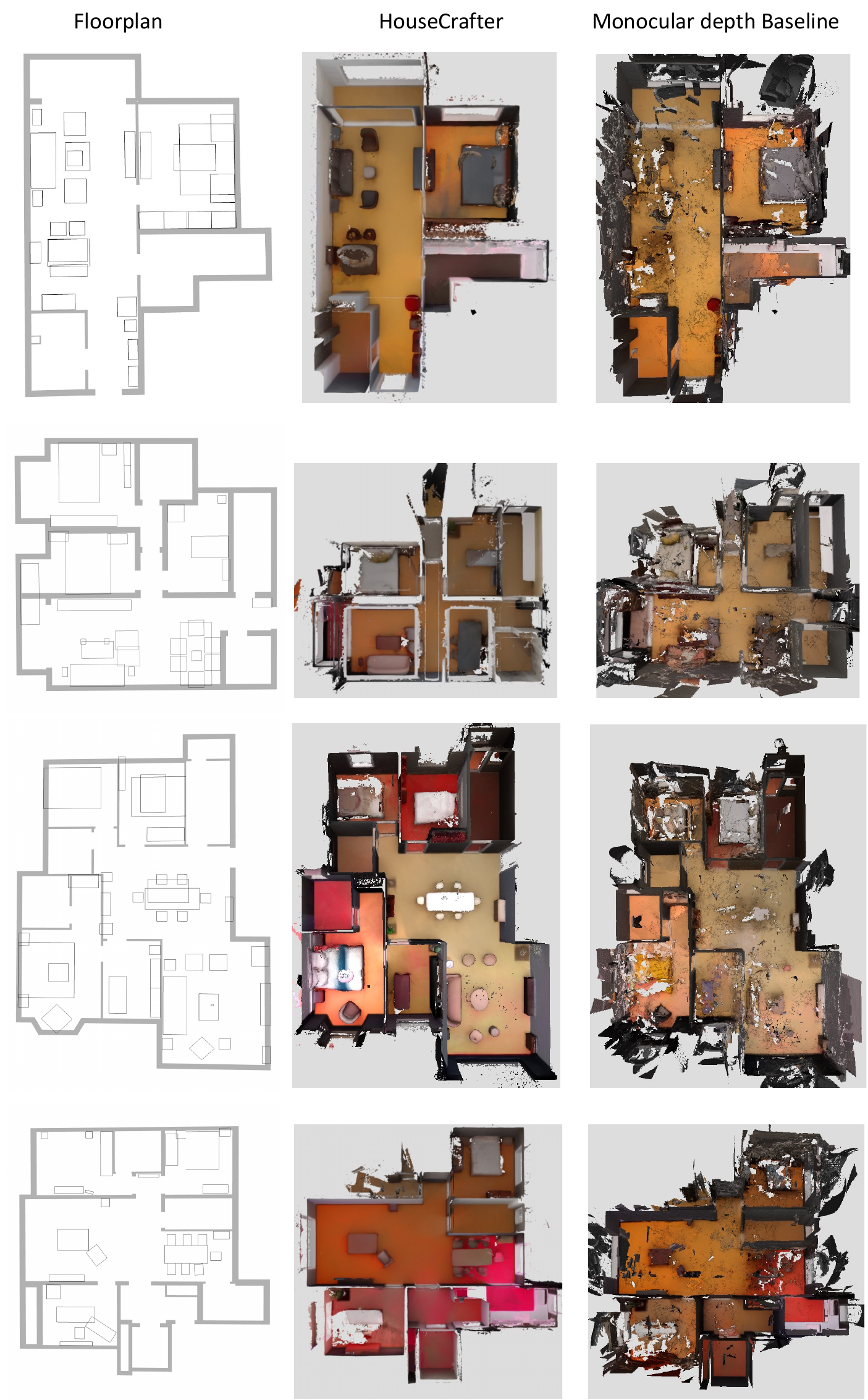}
  \caption{\textcolor{black}{\textbf{Comparison of generated depth with monocular estimated depth.}}}
  \label{fig:monocular_depth}
\end{figure*}

\section{\textcolor{black}{Ablation on Simultaneous RGB and Depth Generation}}
\label{sec:monocular_depth}

\textcolor{black}{
To further show the effectiveness of the RGB-D generation over RGB-only generation in the reconstructed scene, we do an ablation by replacing the generated depth with the estimated depth from an off-the-shelf \emph{metric} monocular depth estimation model~\citep{piccinelli2024unidepth}. Specifically, in each generation batch, we use the monocular estimated depth of the reference RGB images as the reference depths to generate the novel view RGB then estimate the depth for the novel views. As the estimated monocular depth may have an incorrect scale, we use visual cues such as wall or floor to calibrate the scale when possible. The quantitative evaluation in terms of floorplan consistency of the reconstructed scene shows that generating RGB-D images together achieves better results (Table~\ref{tab:monodepth}).
Visualization of the reconstructed scenes  shown in Fig.~\ref{fig:monocular_depth} further confirms the design choice of simultaneous generation of RGB and depth images.
}

\definecolor{mygray}{gray}{0.5}
\renewcommand{\gtres}[1]{\textcolor{mygray}{#1}}

     
\begin{table}[t]

  \caption{\textcolor{black}{\textbf{Ablation study of generating RGB and depth images at the same time.}}
  The results are obtained over \emph{rendered} images from the generated scenes.
  }
  \centering
  \begin{tabular}{lc}
  \toprule
  \label{tab:monodepth}
    \multirow{2}{*}{Method} & Floorplan Consistency \\ 
    \cmidrule(lr){2-2} 
    & mAP@25$\uparrow$ \\ 
    \midrule
     Monocular metric depth &  30.13 \\
     Generated depth \textbf{(proposed)} &  \textbf{45.77} \\
     \midrule
     
     \gtres{GT, 3D-FRONT} & \gtres{54.51} \\ 
    \bottomrule
  \end{tabular}
\end{table}

\section{Ablation Study on Floorplan Encoding}
\label{sec:floor_baseline}
We discuss and ablate an alternative design for floorplan encoding, which does not explicitly use the geometry information. Here, each object in the scene is represented by a vector (encoded with object category and 2D bounding box). 
Each image token/feature at a novel view will then cross-attend to features of all the objects that are within the camera's frustum.
Compared with this alternative design choice, which models \emph{all} the candidate objects within each view, the design presented in the main paper directly considers \textbf{objects on optical rays only} and is able to model the occlusion (\eg, because of walls) better, and thus reduces the number of objects to consider and simplifies the model training.



As shown in Table \ref{tab:ablation_layout}, compared to this baseline, the proposed method in the main paper has higher floorplan consistency (in terms of mAP@25) w.r.t. the input floorplan and higher quality in terms of generated RGB images, validating the effectiveness of our proposed method. 
This baseline design achieves better consistency for depth images. 
We hypothesize that this is due to extra information provided in each object's bounding box instead of the coordinate information of the intersection point. 
Nevertheless, considering the overall quality, we choose the design choice reported in the main paper.

\begin{table*}[t]

  \caption{\textcolor{black}{\textbf{Ablation studies for layout encoding.} The better results are highlighted with \textbf{bold}.}}
  \label{tab:ablation_layout}
  \centering
  \setlength{\tabcolsep}{3.6pt}
  \begin{tabular}{ccccccccccc}

    \toprule
    \multirow{3}{*}{Variant} &  \multicolumn{4}{c}{RGB Metrics} & \multicolumn{4}{c}{Depth Metrics} & \multicolumn{2}{c}{Floor. Const.} \\ 
    \cmidrule(lr){2-5} \cmidrule(lr){6-9} \cmidrule(lr){10-10}
     &    \multirow{2}{*}{FID $\downarrow$} & \multirow{2}{*}{IS $\uparrow$} & \multicolumn{2}{c}{PSNR $\uparrow$} & \multicolumn{2}{c}{AbsRel $\downarrow$} & \multicolumn{2}{c}{$\delta_{0.5}$$\uparrow$} &    \multirow{2}{*}{mAP@25 $\uparrow$} \\
    \cmidrule(lr){4-5}  \cmidrule(lr){6-7}  \cmidrule(lr){8-8} 
     &  &  & R-N & N-N & R-N & N-N & R-N & N-N &&\\
     
    \cmidrule(lr){1-1} \cmidrule(lr){2-5} \cmidrule(lr){6-9} \cmidrule(lr){10-11}
     baseline &27.15& 4.20& 25.01 & \textbf{25.27} & \textbf{4.59} & \textbf{6.89} & \textbf{96.62} & \textbf{93.23} & 38.16\\
     proposed &\textbf{16.70}& \textbf{4.74}& \textbf{25.31} & 24.69 & 6.79 & 7.37 & 92.20 & 92.65 &\textbf{52.26}\\
     \cmidrule(lr){1-1} \cmidrule(lr){2-9} \cmidrule(lr){10-11}
     \gtres{GT}  & \gtres{-} &\gtres{-} & \gtres{-} & \gtres{-} & \gtres{-} & \gtres{-} &\gtres{-}  & \gtres{-}  &\gtres{54.51}\\
    \bottomrule
  \end{tabular}
\end{table*}

\section{Limitations and Future Directions}
\label{sec:limitation}


We show promising results on a challenging task to convert a 2D floorplan into a complete textured 3D scene. Here we briefly discuss the limitations of our approach and dicuss future directions.

First, we seperate the RGB-D image generation and scene reconstruction using TSDF fusion without considering their synergies.
In the future, we aim to explore combining them together, where a single model is sufficient, unifying the generation and reconstruction tasks.

Second, we adopt the TSDF fusion to produce reasonably good results in fusing generated RGB-D images as explicit scene representations are desired to support interactions with the scene. However, the fused scene mesh does not always produce high-quality rendering results. We also experimented with other methods, such as SuGaR~\cite{guedon2023sugar}, which however did not yield better results. An appealing future direction would be to investigate high-quality mesh generation based on NeRF~\cite{mildenhall2021nerf} and 3D Gaussian Splatting~\cite{kerbl3Dgaussians}.



Finally, in our proposed method of injecting the floorplan guidance to the generation process, only the geometry and semantics of the object are leveraged, while the information about the object instance is omitted. We believe that instance-awareness can give better scene understanding thus generating more faithful 3D scenes to the floorplans.

\section{Additional Visual Comparisons}
\label{sec:additional_vis_comp}
\begin{figure*}[h]
  \centering
  \graphicspath{{img/}}
  \includegraphics[width = \linewidth]{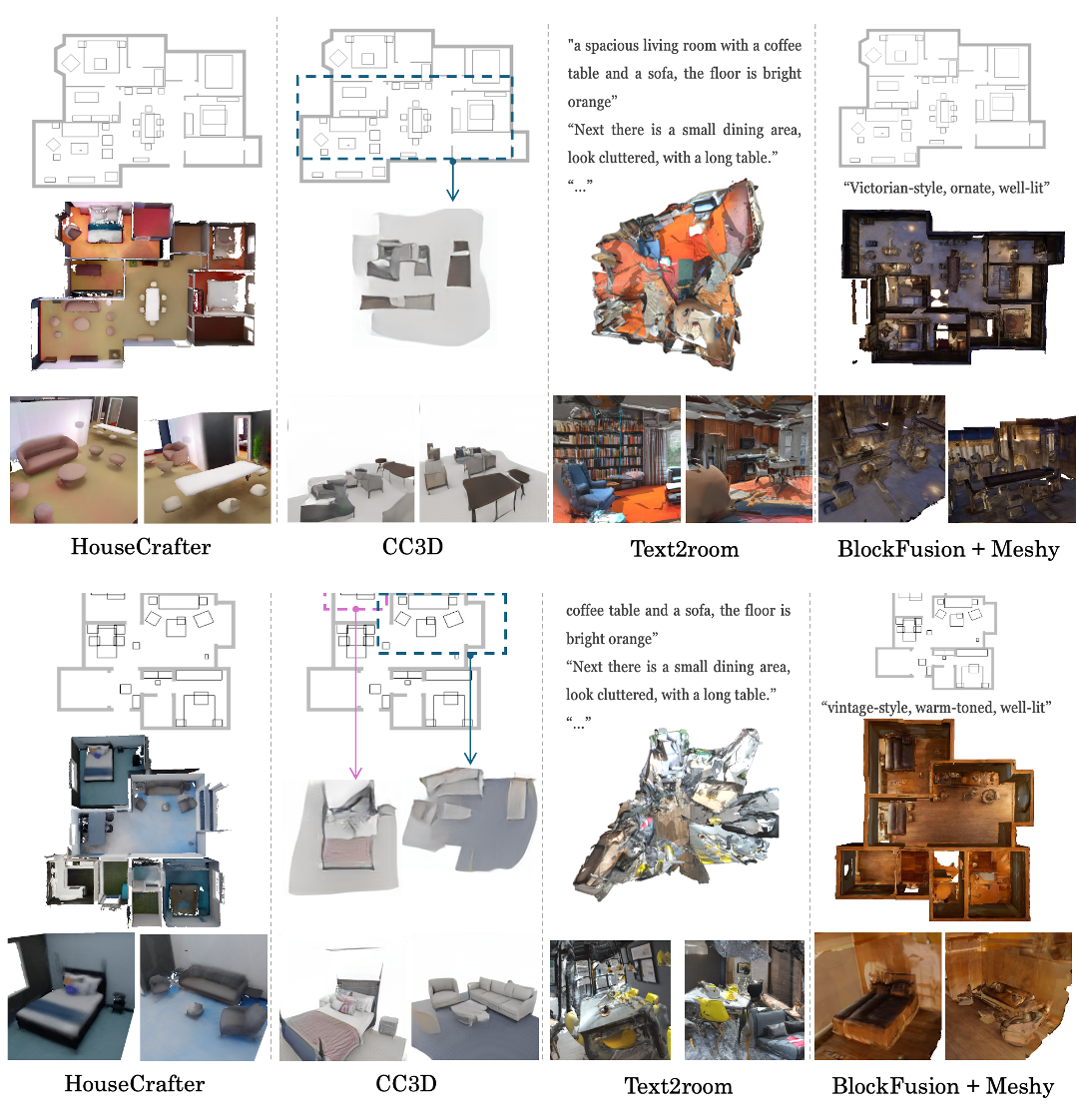}
  \caption{\textbf{Additional comparisons with baseline methods.}}
  \label{fig:extended_comparision}
\end{figure*}

\begin{figure*}[h]
  \centering
  \graphicspath{{img/}}
  \includegraphics[width = \linewidth]{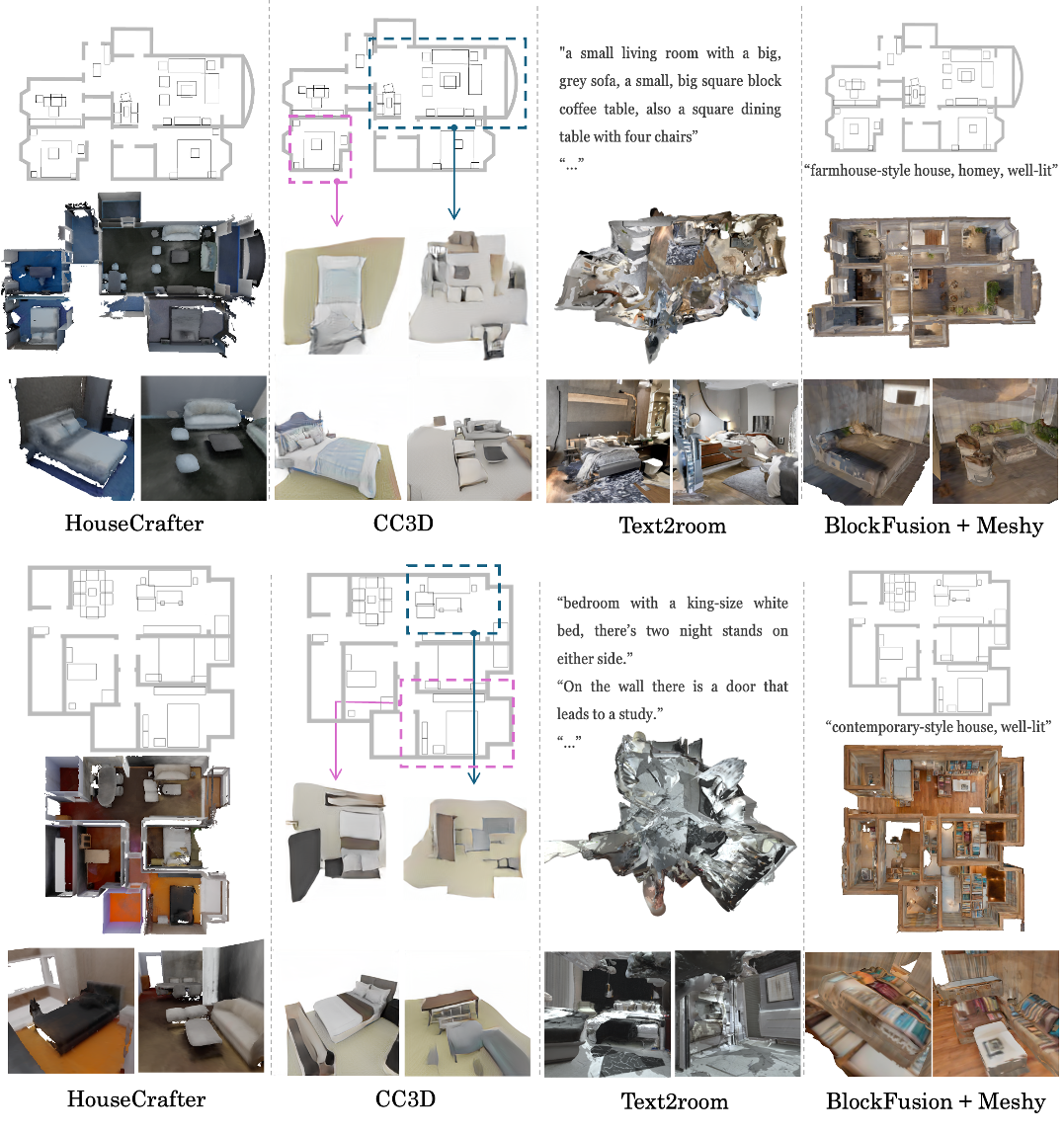}
  \caption{\textcolor{black}{\textbf{Additional comparisons with baseline methods.}}}
  \label{fig:extended_blockfusion_comparison}
\end{figure*}

We show additional visual comparisons with other baseline methods in Fig.~\ref{fig:extended_comparision} and Fig.~\ref{fig:extended_blockfusion_comparison}.
We can see our model produces better scene generation results.
We also refer readers to the supplmentary video for more visual results.

\begin{figure*}[h]
  \centering
  \graphicspath{{img/}}
  \includegraphics[width = \linewidth]{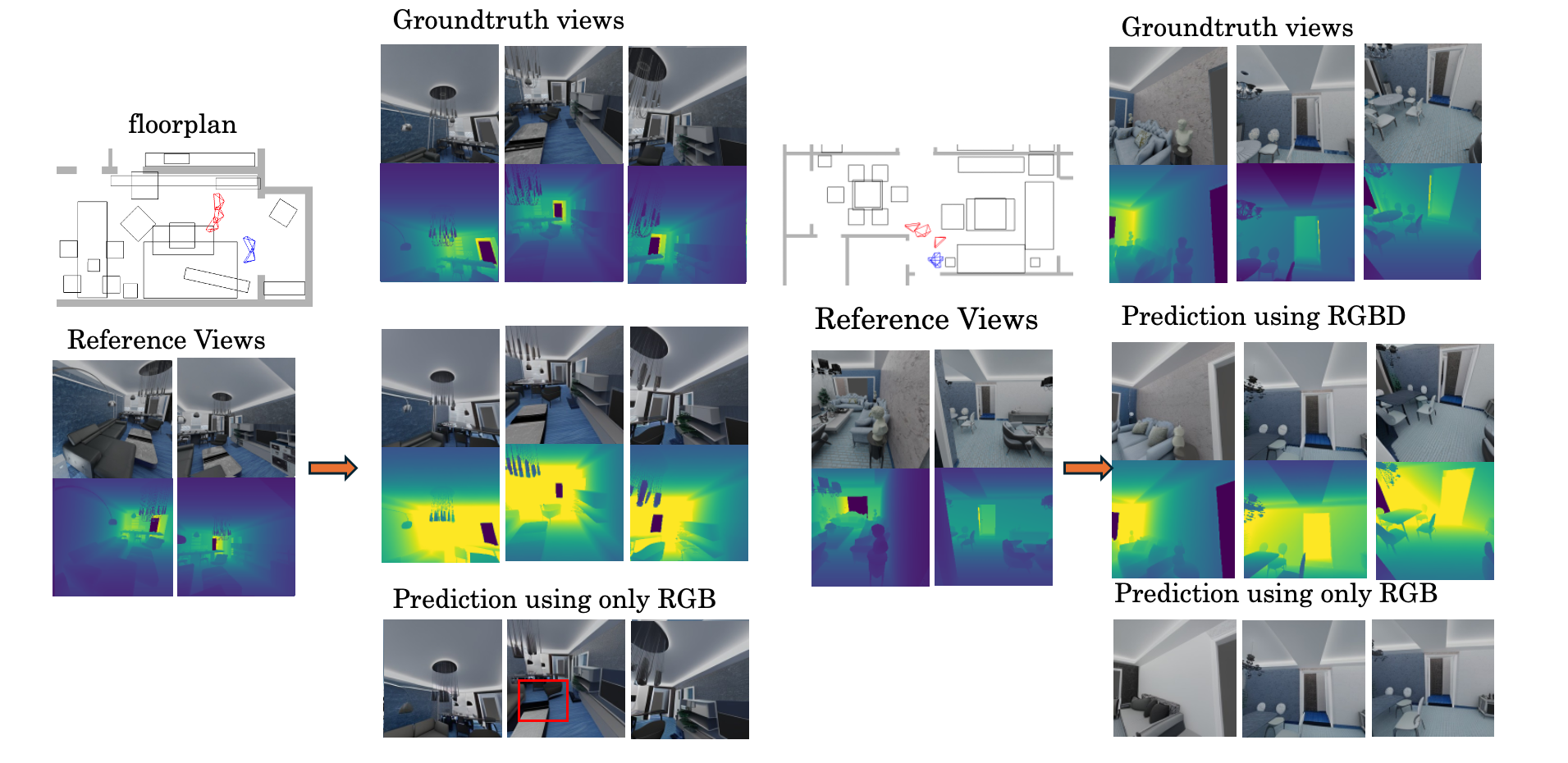}
  \caption{\textbf{Additional visual results of novel view RGB-D image generation.}}
  \label{fig:extended_rgbdnvs_app}
\end{figure*}

\begin{figure*}[h]
  \centering
  \graphicspath{{img/}}
  \includegraphics[width = \linewidth]{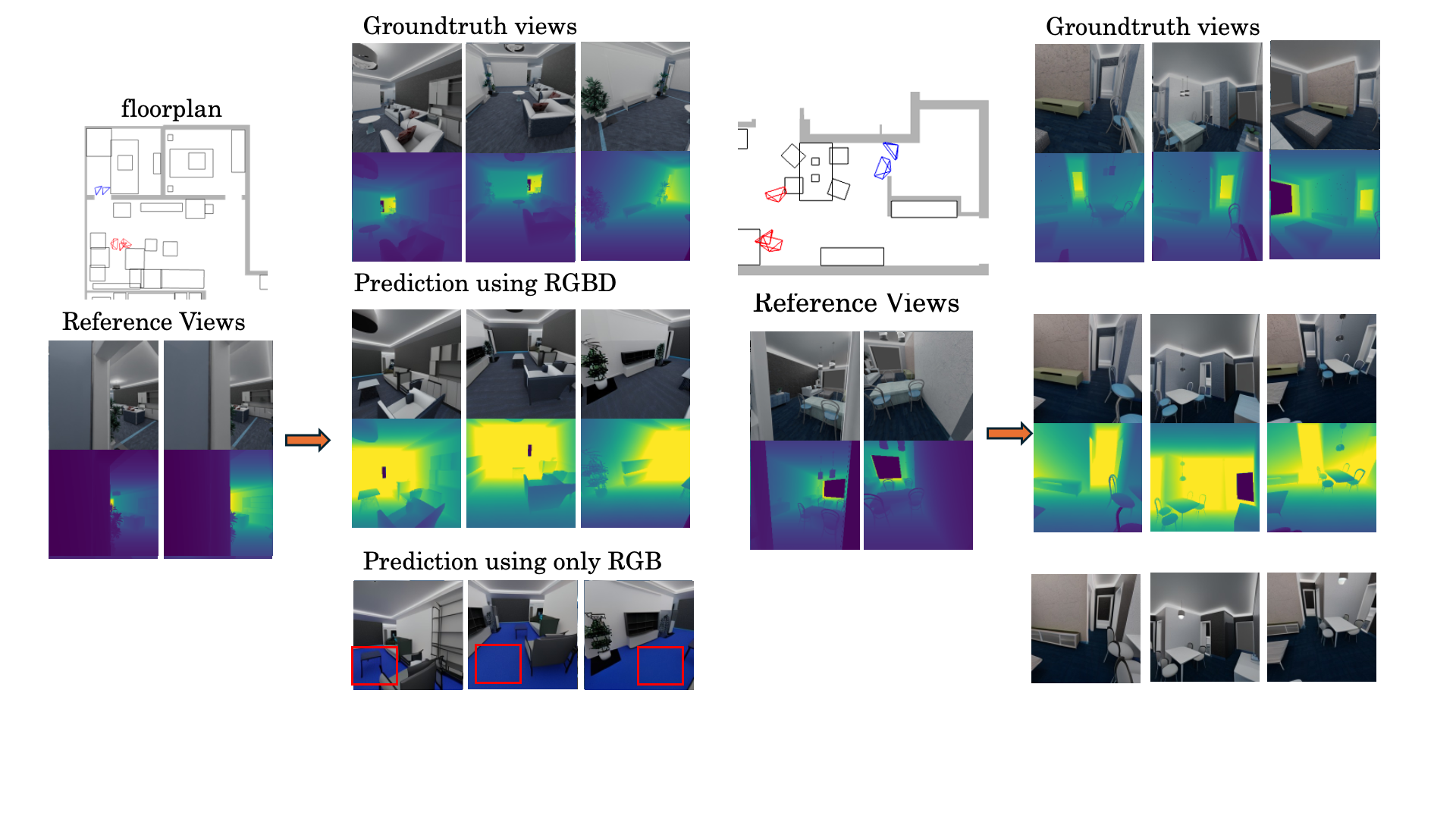}
  \caption{\textcolor{black}{\textbf{Additional visual results of novel view RGB-D image generation.}}}
  \label{fig:extended_rgbdnvs}
\end{figure*}

\section{Generalization to Real-World Data}

\textcolor{black}{Our primary experiments are conducted on the synthetic 3D-FRONT dataset, which allows us to generate high-quality multi-view training data and to define generation trajectories with full control. However, extending to real-world data poses additional challenges due to noisy depth measurements and constrained, predefined camera trajectories.}

\textcolor{black}{To evaluate the model's generalization capabilities in real-world scenarios, we adopt the recently released Stable Virtual Camera (SEVA) framework~\citep{seva} as the backbone and integrate it with our proposed depth and layout conditioning modules.}

\textcolor{black}{For RGBD generation, we follow the same channel expansion strategy applied in our Stable Diffusion 1.5 experiments, modifying SEVA's input and output layers accordingly. Additionally, we revise the Plücker ray encoding scheme to incorporate both depth-projected 3D positions and ray directions more effectively.}

\textcolor{black}{For layout conditioning, we insert our layout attention modules at the end of each ResNet block within the UNet architecture, enabling the model to incorporate high-level spatial priors during generation.}

\begin{figure*}[h]
  \centering
  \includegraphics[width=0.95\linewidth]{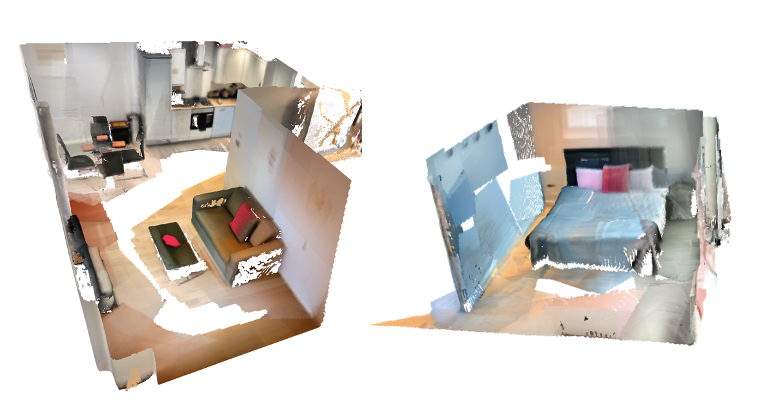}
  \vspace{-6pt}
   \caption{\textbf{Generation results of our model on ArkitScenes.}}
   \vspace{-10pt}
   \label{fig:arkitscene_vis}
\end{figure*}

\textcolor{black}{We fine-tune the resulting model on the ARKitScenes~\citep{arkitscenes} dataset and present qualitative results in Fig.\ref{fig:arkitscene_vis} to demonstrate its performance and generalization potential in real-world settings.}

\begin{table}[!t]
  \caption{\textbf{Quantitative measurement of our result on ArkitScenes data}
Similar to the main paper, the scene quality is measured in Inception Score(IS), Floorplan Consistency, and 3D quality}
  \label{depth2d_app}
  \centering
  \small
  \setlength{\tabcolsep}{1.5pt}

  \begin{tabular}{ccccl}
  \toprule
  \label{tab:baseline_app}
  \multirow{2}{*}{Method} & Visual Quality & Floorplan Consistency  &3D quality\\ 
    \cmidrule{2-4}
    & IS $\uparrow$ & mAP@25$\uparrow$  & artifacts$\downarrow$\\ 
    \midrule
    Text2Room & \textbf{5.35}& -  &302.5\\
     \textbf{HouseCrafter}(realworld) & 5.57 & \textbf{49.76}    & 224.0\\
    \bottomrule
  \end{tabular}
  \vspace{-15pt}
\end{table}
\section{Ablation on the number of reference and target views.}
\begin{figure*}[h]
  \centering
  \includegraphics[width=0.95\linewidth]{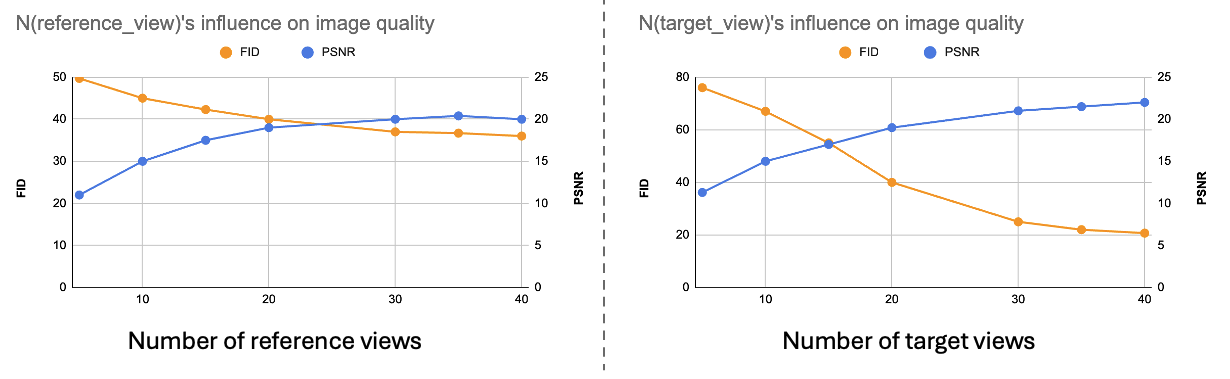}
   \caption{\textbf{Influence of selected reference view and target view numbers, respectively.}}
   \label{fig:influence}
\end{figure*}

\textcolor{black}{In average, we select 30 reference and 10 target views, respectively, at each location for a single house generation. 
Here, we analyze how the performance varies with different numbers of reference and target views. In this experiment, we re-grouped the location and views, so that as the reference and target views in each location grow, the locations we need to traverse decrease.
As can be seen in the Fig.\ref{fig:influence} above, the generation quality improves when more reference and target views are involved in a location (thus fewer locations) and starts saturating at a certain threshold. 
Due to the GPU memory constraints, we can not use significantly large number of reference and target views.}


